\documentclass[fleqn,10pt]{wlscirep}
\usepackage[utf8]{inputenc}
\usepackage[T1]{fontenc}
\usepackage{graphicx}%
\usepackage{subfigure}%
\usepackage{amsmath,amssymb,amsfonts}%
\usepackage{amsthm}%
\usepackage{amssymb}
\usepackage{mathtools}
\usepackage[ruled,norelsize]{algorithm2e}
\usepackage{algorithmic}
\usepackage[makeroom]{cancel}
\usepackage{array}
\usepackage[title]{appendix}%
\usepackage{xcolor}%
\usepackage{textcomp}%
\usepackage{verbatim}%
\usepackage{manyfoot}%
\usepackage{booktabs}%
\usepackage{multirow}%
\usepackage{listings}%
\usepackage{float}
\usepackage[absolute]{textpos}

\newtheorem{theorem}{Theorem}
\newtheorem{remark}{Remark}%

\newtheorem{prop}{Proposition} 
\newtheorem{lemma}[theorem]{Lemma}

\title{Relative Importance Sampling for off-Policy Actor-Critic in Deep Reinforcement Learning}

\author[1,2,3,*]{Mahammad Humayoo}
\author[4]{Gengzhong Zheng}
\author[4]{Xiaoqing Dong}
\author[3]{Liming Miao}
\author[3]{Shuwei Qiu}
\author[3]{Zexun Zhou}
\author[3]{Peitao Wang}
\author[2,5]{Zakir Ullah}
\author[3]{Naveed Ur Rehman Junejo}
\author[1,2]{Xueqi Cheng}

\affil[1]{CAS Key Laboratory of Network Data Science and Technology, Institute of Computing Technology, CAS, Beijing, 100190, China}
\affil[2]{University of Chinese Academy of Sciences, Beijing, 101408, China}

\affil[3]{School of Computer and Information Engineering, Hanshan Normal University, Guangdong, 521041, China}

\affil[4]{School of Physics and Electronic Engineering, Hanshan Normal University, Guangdong, 521000, China}

\affil[5]{School of Data Science, Capital University of Economics and Business, Beijing, 100070, China}

\affil[*]{humayoo@hstc.edu.cn}


\begin{abstract}
Off-policy learning exhibits greater instability when compared to on-policy learning in reinforcement learning (RL). The difference in probability distribution between the target policy ($\pi$) and the behavior policy (b) is a major cause of instability. High variance also originates from distributional mismatch. The variation between the target policy's distribution and the behavior policy's distribution can be reduced using importance sampling (IS). However, importance sampling has high variance, which is exacerbated in sequential scenarios. We propose a smooth form of importance sampling, specifically relative importance sampling (RIS), which mitigates variance and stabilizes learning. To control variance, we alter the value of the smoothness parameter $\beta\in[0, 1]$ in RIS. We develop the first model-free relative importance sampling off-policy actor-critic (RIS-off-PAC) algorithms in RL using this strategy. Our method uses a network to generate the target policy (actor) and evaluate the current policy ($\pi$) using a value function (critic) based on behavior policy samples. Our algorithms are trained using behavior policy action values in the reward function, not target policy ones. Both the actor and critic are trained using deep neural networks. Our methods performed better than or equal to several state-of-the-art RL benchmarks on OpenAI Gym challenges and synthetic datasets.
\end{abstract}
\begin{document}

\flushbottom
\maketitle
%
%
\thispagestyle{empty}

\begin{textblock}{100}(1.5,0.5)
\noindent\Large Published in Nature Scientific Reports. Access it via the following URL:\newline \url{https://www.nature.com/articles/s41598-025-96201-5}
\end{textblock}

\section{Introduction}
\label{introduction}
Various intricate challenges have been tackled using model-free deep RL methods \cite{Sutton2018Reinforcement,Silver2016MasteringTG,Silver2017MasteringTG,Mnih2013PlayingAW,Mnih2016Asynchronous,Schulman2015TrustRP,Lillicrap2015Continuous,Gu2016QPropSP}. Model-free RL learning encompasses both on-policy and off-policy approaches. Off-policy approaches enable the simultaneous learning of a target policy while observing and gathering data from another policy, known as the behavior policy. It means that an agent learns about a policy distinct from the one it is carrying out while there is a single policy (i.e., target policy) in on-policy methods. It means that the agent learns only about the policy it is carrying out. Simply put, if two policies are identical (i.e., $\pi=b$), then the arrangement is referred to as on-policy. Alternatively, the scenario is referred to as off-policy, if $\pi$ is not equal to $b$ \cite{harutyunyan2016q,Degris2012Off,precup2001off,Gu2016QPropSP,Hanna2018ImportanceSP,gruslys2017reactor}.

\begin{figure}[!htbp]
\subfigure[The off-policy learning.]{
\includegraphics[width=0.5\columnwidth]{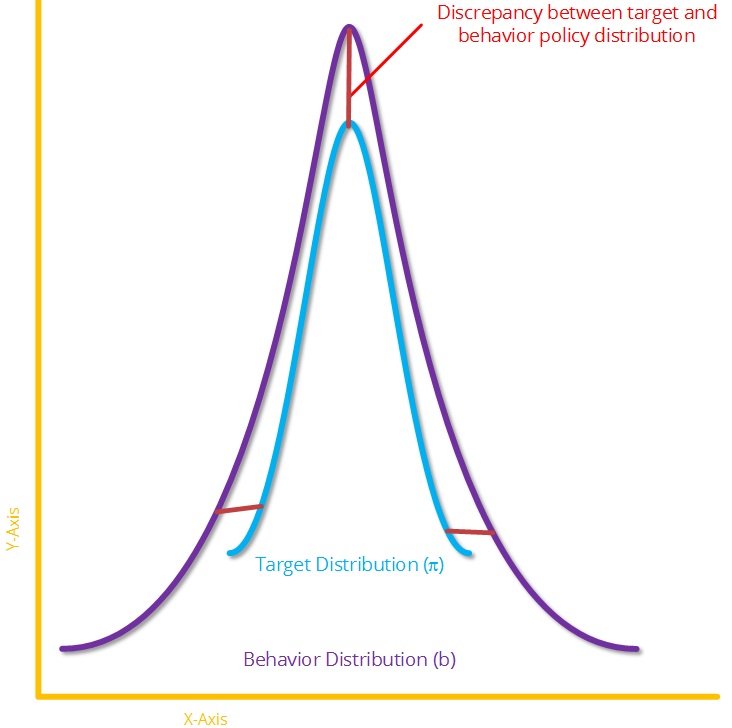}
\label{fig:offpolicy_dist}
}%
\subfigure[The on-policy learning.]{
\includegraphics[width=0.5\columnwidth]{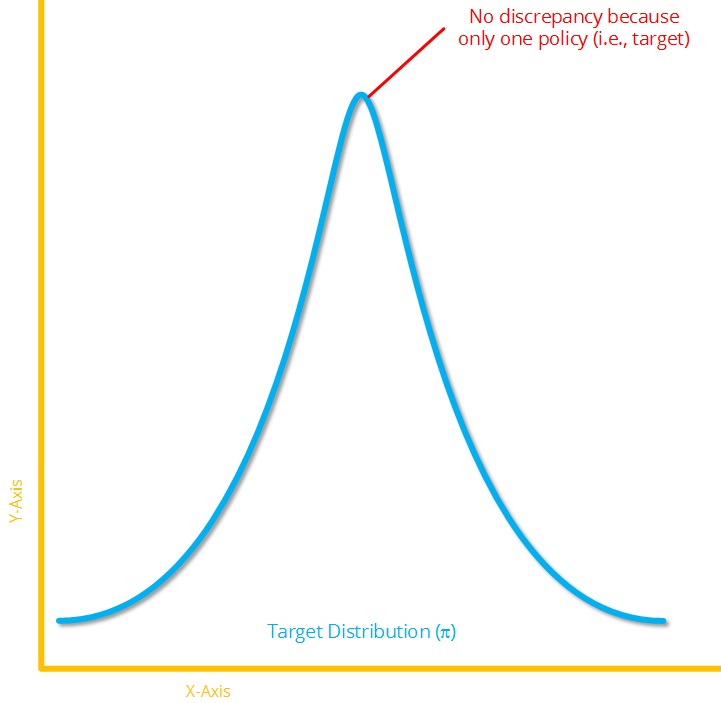}
\label{fig:onpolicy_dist}
}
\caption{A comparison of on- and off-policy learning.}
\end{figure}

Fig.\ref{fig:offpolicy_dist} illustrates that off-policy learning primarily involves two policies: the behavioral policy (b), also known as the sampling distribution, and the target policy ($\pi$), also known as the target distribution. The Fig.\ref{fig:offpolicy_dist} also shows that there is often a discrepancy between these two policies ($\pi$ and b). This discrepancy makes off-policy unstable and introduces significant variance \cite{precup2000eligibility,levine2020offline, Munos2016SafeAE,thomas2015high,fujimoto2019off,nachum2019algaedice, jiang2016doubly}; a bigger difference between these policies, instability is also high, and a smaller difference between these policies, instability is also low in off-policy learning, whereas on-policy has a single policy (i.e., target policy), as shown in Fig.\ref{fig:onpolicy_dist}. The instability is not an issue for on-policy learning due to the sole policy. Therefore, compared to off-policy, on-policy is more stable.\par

In addition to the aforementioned benefits, there are other advantages and disadvantages associated with off-policy and on-policy learning. On-policy approaches, while unbiased, frequently encounter challenges like sample inefficiency. Off-policy approaches are characterized by higher sampling efficiency and are safe, yet they may exhibit instability and add variance. Both on-policy and off-policy approaches have their limitations. Consequently, multiple approaches have been suggested to address the shortcomings of each strategy. For instance, it is possible for on-policy methods to attain a comparable level of sample efficiency as off-policy methods \cite{Gu2016QPropSP,Mnih2016Asynchronous,Schaul2015PrioritizedER,Schulman2015TrustRP,van2014off}. Similarly, off-policy methods can achieve a similar level of stability as on-policy methods \cite{Degris2012Off,Mahmood2014Weighted,gruslys2017reactor,Wang2016SampleEA,Haarnoja2018SoftAO} and mitigate the variance induced by distributional mismatch \cite{jiang2016doubly,kallus2020statistically}.\par
In practical reinforcement learning (RL) contexts, such as autonomous driving or robotic control, the policies that provide data frequently diverge substantially from the target policies. This distributional discrepancy might result in instability and elevated variance throughout the learning process, especially in continuous action spaces \cite{precup2000eligibility,Munos2016SafeAE,jiang2016doubly}. A key motivation for this study to mitigate instability and variance in off-policy learning. Recent improvements have underscored the advantages of divergent behaviour policy for exploration; nonetheless, these methods are frequently restricted in long-horizon tasks or offline reinforcement learning contexts, where exploration is restricted or impractical. Furthermore, numerous approaches depend on rigid assumptions about the behaviour policy or reward structure, hence limiting their application in intricate, real-world environments \cite{yarats2021reinforcement,levine2020offline}. The suggested RIS-off-PAC algorithm seeks to address these constraints by dynamically adjusting for distribution discrepancies via relative importance sampling. This method guarantees a consistent learning process. RIS-off-PAC enhances reliability and scalability in off-policy learning by reducing instability and variance, which is crucial for real-world applications when data acquisition is costly or limited.\par

Importance sampling is a well-known method to evaluate off-policy, permitting off-policy data to be used as if it was on-policy \cite{Hanna2018ImportanceSP}. IS can be used to study one distribution while a sample is made from another distribution \cite{owen2013monte}. The degree of deviation of the target policy from the behavior policy at each time t is captured by the importance sampling ratio (i.e., $IS = \frac{\pi(A_{t}|S_{t})}{b(A_{t}|S_{t})}$) \cite{precup2001off}. IS is also considered as a technique for mitigating the variance of the estimate of an expectation by cautiously determining sampling distribution (b). Our new estimate has low variance, if b is chosen properly. The variance of an estimator relies on how much the sampling distribution and the target distribution are unlike \cite{rubinstein2016simulation}. For theory behind importance sampling that is presented here, we refer to see \cite[Chapter 9]{owen2013monte} for more details.\par

An additional factor contributing to the instability of off-policy learning is the lack of uniformity in the values generated by importance sampling (IS) for different samples. The IS occasionally produces a high value for certain samples and a low value for other samples, hence amplifying the disparity between the two distributions. Authors\cite{Yamada2011RelativeDE} introduced a smooth version of importance sampling called the relative importance sample. This method was proposed to address the instability in semi-supervised learning. We apply this technique in deep reinforcement learning to alleviate the discrepancy between $\pi$ and b, hence diminishing the variation and instability associated with off-policy learning. Some notable methods based on Importance Sampling (IS) include: Weighted Importance Sampling (WIS) \cite{Mahmood2014Weighted}, Sample Efficient Actor-Critic with Experience Replay (ACER) \cite{Wang2016SampleEA}, Retrace \cite{Munos2016SafeAE}, Q-prop \cite{Gu2016QPropSP}, Soft Actor-Critic (SAC) \cite{Haarnoja2018SoftAO}, Off-Policy Actor-Critic (Off-PAC) \cite{Degris2012Off}, The Reactor \cite{gruslys2017reactor}, Guided Policy Search (GPS) \cite{Levine2013GuidedPS}, Efficient Multiple Importance Sampling (MIS) \cite{elvira2015efficient}, and others.\par

In summary, the following contributions are made by this paper:
(i) We develop a simple Relative Importance Sampling (RIS) estimator that improves stability and diminishes the variance of off-policy approaches. (ii) We provide an off-policy actor-critic method, termed RIS-off-PAC, that utilises relative importance sampling in deep reinforcement learning. As far as we know, we are presenting the first instance of RIS with actor-critic. Furthermore, we investigate a variation of the actor-critic method termed the natural gradient actor-critic, which employs relative importance sampling. This form, known as the relative importance sampling-off-policy natural actor-critic (RIS-off-PNAC), substantially enhances our contributions. (iii) On benchmark problems, The RIS estimator exhibits performance that is either superior to or competitive with various state-of-the-art RL benchmarks, while maintaining stable learning.

The remaining sections of the paper are organized as follows: The discussion of related works can be found in section \ref{relatedwork}. In section \ref{preliminaries}, we provide a preliminary. Sections \ref{IS} and \ref{actor-critic} demonstrate the concepts of relative importance sampling and the actor-critic model, respectively. Section \ref{experiment} provides a detailed account of the conducted experiments. Ultimately, we provide a conclusion in section \ref{conclusion}.

\section{Related Work}
\label{relatedwork}
\subsection{On-Policy}
The authors in this study\cite{Thomas2014BiasIN} claimed that biased discounted reward made natural actor-critic algorithms unbiased average reward natural actor-critics. Bhatnagar et al.\cite{Bhatnagar2009NaturalAA} introduced four novel online actor-critic RL algorithms that utilize natural-gradient, function-approximation, and temporal difference learning techniques. In addition, they showcased the convergence of these four algorithms to a local maximum. Schaul et al.\cite{Schaul2015PrioritizedER} presented a paradigm that prioritizes experience, allowing for more frequent replay of important transitions, resulting in more efficient learning. When the standard Gaussian distribution was employed as a stochastic policy, the presence of bounded actions resulted in bias. Chou et al.\cite{Chou2017ImprovingSP} proposed the utilization of the beta distribution as an alternative to the Gaussian distribution. They investigated the balance between bias and variance of the policy gradient for both on-policy and off-policy scenarios.\par

Mnih et al.\cite{Mnih2016Asynchronous} introduced four asynchronous deep RL methods in their study. The most efficient approach was the asynchronous advantage actor-critic (A3C) algorithm, which involved maintaining a policy $\pi(a_{t}|s_{t};\theta)$ and an estimated value function $V(s_{t};\theta_{v})$. Van Seijen and Sutton\cite{Van2014True} proposed a genuine online TD($\lambda$) learning method, which is similar to an online forward view. This algorithm demonstrated superior performance compared to its traditional counterpart in both prediction and control tasks. Schulman et al.\cite{Schulman2015TrustRP} devised an approach known as Trust Region Policy Optimization (TRPO) that delivers policy improvements in a monotonic manner. They also generated a practical algorithm that exhibits superior sample efficiency and performance. Schulman et al.\cite{Schulman2015HighDimensionalCC} introduced a technique called generalized advantage estimation (GAE) to reduce variance in policy gradient. This method utilizes a trust region optimization approach for the value function. The GAE policy gradient effectively reduced variation while preserving an acceptable amount of bias. Our focus lies on off-policy learning as opposed to on-policy learning.
\subsection{Off-Policy}
Yarats et al.\cite{yarats2021reinforcement} proposed Proto-RL framework that has highlighted the advantages of divergent behaviour policies for exploration, especially in environments with sparse rewards. However, Proto-RL may encounter difficulties in environments with extensive or continuous action spaces, necessitating more sophisticated or diverse exploration strategies. RIS-off-PAC tackles the significant challenge of high variance resulting from distribution mismatch, a concern that is particularly evident in continuous action spaces, especially in unrestricted environments with rewards. Levine et al.\cite{levine2020offline} examine the difficulties associated with distribution mismatch in offline RL and its impact on the stability of learning systems, particularly when utilising off-policy data. Hachiya et al.\cite{hachiya2009adaptive} examined the variance of the value function estimator in off-policy approaches to manage the balance between bias and variance. Mahmood et al.\cite{Mahmood2014Weighted} employed weighted importance sampling in combination with function approximation to develop a novel variant of off-policy LSTD($\lambda$) known as WIS-LSTD($\lambda$). Degris et al.\cite{Degris2012Off} introduced the off-policy actor-critic (off-PAC) technique, where an agent acquires samples from a behavior policy while learning a target policy. Gruslys et al.\cite{gruslys2017reactor} introduced a RL agent called Reactor, which is efficient in terms of sample usage and utilizes an actor-critic approach. The critic was trained using the off-policy multi-step Retrace technique, while the actor was trained using a new policy gradient approach termed B-leave-one-out. Zimmer et al.\cite{zimmer2016off} presented a novel off-policy actor-critic RL system that addresses the challenge of continuous state and action spaces by leveraging neural network techniques. Their approach also enabled the balancing of data-efficiency and scalability. Levine and Koltun \cite{Levine2013GuidedPS} discussed the use of "guided policy search" (GPS) to prevent the occurrence of "poor local optima" in intricate policies that involve numerous variables. GPS utilized "differential dynamic" programming to generate suitable guiding samples and formulated a "regularized importance sampled policy optimization" that incorporated these samples into policy exploration.\par

Lillicrap et al.\cite{Lillicrap2015Continuous} proposed a method called deep deterministic policy gradient (DDPG) that uses deep function approximators and the deterministic policy gradient (DPG) to learn policies in continuous action spaces. This algorithm is model-free and off-policy. In their study, Wang et al.\cite{Wang2016SampleEA} introduced a robust and efficient actor-critic deep RL agent named ACER. This agent incorporates "experience replay" and is capable of effectively handling both continuous and discrete action spaces.  ACER employed the techniques of "truncate importance sampling with bias correction, stochastic dueling network architectures, and efficient trust region policy optimization" to accomplish its goal. Munos et al.\cite{Munos2016SafeAE} introduced a new approach, named Retrace($\lambda$), which possesses three key characteristics: low variance, safety through the utilization of samples obtained from any behavior policy, and efficiency in estimating the Q-Function from off-policy data. Gu et al.\cite{Gu2016QPropSP} introduced a technique named Q-Prop, which demonstrated high efficiency in terms of sample usage and stability. It combined the benefits of on-policy methods (policy gradient stability) and off-policy methods (efficiency). Model-free deep RL learning algorithms commonly face two primary challenges: significant sampling inefficiency and instability. Haarnoja et al.\cite{Haarnoja2018SoftAO} introduced a soft actor-critic (SAC) approach in their work, which utilizes maximum entropy and off-policy techniques. Off-policy ensured efficient use of given samples, while entropy maximization ensured stability. The majority of these methods employ either the traditional IS or entropy method, while we utilize the RIS method. To obtain a comprehensive review of the IS-off-Policy technique, refer to the publications by \cite{precup2000eligibility,sutton2016emphatic,jie2010connection,elvira2015efficient,gu2017interpolated,van2014off,precup2001off}.

\section{Preliminaries}
\label{preliminaries}
A Markov decision process (MDP) is a mathematical model used to represent problems in the field of RL. A MDP is characterized by a set of items, represented by the tuple ($\mathcal{S}$, $\mathcal{A}$, $\mathcal{R}$, $\mathbb{P}$, $\gamma$). The set $\mathcal{S}$ represents the possible states, $\mathcal{A}$ represents the possible actions, $\mathcal{R}$ represents the distribution of rewards for each (state, action) pair, $\mathbb{P}$ represents the transition probability (i.e., the distribution of the next state given a (state, action) pair), and $\gamma\in(0, 1]$ represents a discount factor. The symbols $\pi$ and b represent the target policy and behavior policy, respectively. A policy ($\pi$) is a mapping between the set of states ($\mathcal{S}$) to the set of actions ($\mathcal{A}$), which determines the action to be taken in each state. In classical RL, an agent engages with an environment through a series of distinct time intervals. At each time step t, the agent selects an action $a_{t}\in\mathcal{A}$ based on its policy ($\pi$) and the current state $s_{t}\in\mathcal{S}$. As a result, the agent receives the subsequent state $s_{t+1}\in\mathcal{S}$ based on the transition probability $\mathbb{P}(s_{t+1}|s_{t},a_{t})$ and perceives a single numerical reward $r_{t}(s_{t},a_{t})\in\mathcal{R}$. The procedure continues until the agent reaches the terminal state, at which point the process restarts. The agent outputs $\gamma$-discounted total accumulated return from each state $s_{t}$ i.e. $R_{t}=\substack{\infty \\ \sum \\k\geq0}\gamma^{k}r(s_{t+k},a_{t+k})$.\\
In the field of RL, there are two common functions used to determine the action to be taken based on a given policy ($\pi$ or b): the state-action value function ($Q^{\pi}(s_{t},a_{t})= \mathbb{E}_{s_{t+1:\infty},a_{t+1:\infty}}[R_{t}|s_{t},a_{t}]$) and the state value function ($V^{\pi}(s_{t})=\mathbb{E}_{a_{t}\in \mathcal{A}}[Q^{\pi}(s_{t},a_{t})]$). $\mathbb{E}$ represents the mathematical concept of expectation, which is also known as the mean. The agent's objective is to maximize the expected return ($J(\theta)= \mathbb{E}_{\pi}[R_{\theta}]$) by employing policy gradient ($\nabla_{\theta}J(\theta)$) with respect to parameter $\theta$. $J(\theta)$ is commonly referred to as an objective or a loss function. The policy gradient of the objective function, as described in notation \cite{Sutton1999PolicyGM}, is denoted as \cite{Schulman2015HighDimensionalCC}, is given by:
\vspace{-2mm}
\begingroup
\allowdisplaybreaks
\begin{align}
\label{pg1}
\nabla_{\theta}J(\theta) = \mathbb{E}_{s_{0:\infty},a_{0:\infty}} \Bigg[\sum\limits_{\substack{t\geq0}} A^{\pi}(s_{t},a_{t})\ \nabla_{\theta}\  log\ \pi_{\theta}(a_{t}|s_{t}) \Bigg]
\end{align}%
\endgroup
The term $A^{\pi}(s_{t},a_{t})$ refers to an advantage function. The authors in this study \cite{Schulman2015HighDimensionalCC} demonstrated that it is possible to substitute several expressions for $A^{\pi}(s_{t},a_{t})$ without introducing bias. These alternatives include the state-action value ($Q^{\pi}(s_{t},a_{t})$), the discounted return $R_{t}$, or the temporal difference (TD) residual ($r_{t}+\gamma V^{\pi}(s_{t+1}) - V^{\pi}(s_{t})$). We incorporate TD residual in our approach. The policy gradient approximator with $R_{t}$ exhibits high variance and low bias, while the approximator utilizing function approximation demonstrates high bias and low variance \cite{Wang2016SampleEA}. IS typically exhibits low bias but large variance, as indicated by multiple sources \cite{sutton2016emphatic,hachiya2009adaptive,Mahmood2014Weighted}. We employ RIS as a substitute for IS. Integrating the advantage function with function approximation and RIS to establish a stable off-policy in RL. Policy gradient with function approximation refers to an actor-critic algorithm \cite{Sutton1999PolicyGM} that optimizes the policy based on feedback from the critic, such as the deterministic policy gradient \cite{Silver2014DeterministicPG,Lillicrap2015Continuous}.

\section{Standard Importance Sampling}
\label{IS}
An inherent cause of instability in off-policy learning is the disparity between distributions. In off-policy RL, our objective is to collect data samples from the target policy distribution. However, in reality, the data samples are obtained from the behavior policy distribution. Importance sampling is a widely recognized method for addressing this type of discrepancy \cite{rubinstein2016simulation,precup2000eligibility}. To illustrate, our objective is to approximate the anticipated value of an action (a) in a given state (s) using samples obtained from the target policy ($\pi$) distribution. However, in actuality, the samples are derived from a different distribution known as the behavior policy (b). One can describe a classical form of importance sampling as:

\begin{align}
\label{is2}
\mu = \mathbb{E}_{\pi}\{R(s,a)\} &= \sum\limits_{\substack{a\sim\pi}}\pi(a|s)R(s,a) \\
                                &= \sum\limits_{\substack{a\sim\pi}}\frac{\pi(a|s)}{b(a|s)}b(a|s)R(s,a) \nonumber \\
                                &= \mathbb{E}_{a\sim b}\left\{\frac{\pi(a|s)}{b(a|s)}R(s,a)\right\}\nonumber
\end{align}

The importance sampling estimate of $\mu$ = $\mathbb{E}_{\pi}\{R(s,a)\}$ is

\begin{align}
\label{is3}
\hat{\mu}_{b} \approx \frac{1}{n}\sum\limits_{\substack{t=1,a\sim b}}^n \frac{\pi(a_{t}|s_{t})}{b(a_{t}|s_{t})}R(s_{t},a_{t})
\end{align}

Where R(s,a) is a discounted reward function, $(s_{t},a_{t})$ are samples drawn from b and IS estimator ($\hat{\mu}_{b}$) computes an average of sample values.

\subsection{Relative Importance Sampling}
While there have been some studies \cite{Wang2016SampleEA,precup2001off,Gu2016QPropSP} conducted on addressing instability, no papers have been discovered that utilize a smooth version of importance sampling in RL. The utilization of the smooth version of IS, such as RIS, serves the purpose of mitigating the instability in semi-supervised learning. The term "quasi RIS" can be defined as:
\begin{align}%
\label{ris4}
\mu_{\beta} = \frac{e^{\pi(a|s)}}{\beta e^{\pi(a|s)} + (1-\beta)e^{b(a|s)}}
\end{align}
This is one of the main contributions of this study. We use RIS in place of classical IS in our method. Then the RIS estimate of $\mu_{\beta}$ = $\mathbb{E}_{\pi}\{R(a|s)\}$ is
\begin{align}
\label{ris5}
\hat{\mu}_{\beta} \approx \frac{1}{n}\sum\limits_{\substack{t\geq0,a\sim b}}^n \frac{e^{\pi(a_{t}|s_{t})}}{\beta e^{\pi(a_{t}|s_{t})} + (1-\beta)e^{b(a_{t}|s_{t})}} R(a_{t}|s_{t})
\end{align}%
\begin{prop}%
\label{prop1}
Since the importance is always non-negative, the relative importance is no greater than $\frac{1}{\beta}$:
\begin{align}
\label{prop6}
\mu_{\beta} = \frac{1}{\beta + (1-\beta)\frac{e^{b(a|s)}}{e^{\pi(a|s)}}} \leq \frac{1}{\beta}
\end{align}
The proof is presented in appendix \ref{proofs}.
\end{prop}%

\section{RIS-off-PAC Algorithm}
\label{actor-critic}
An actor-critic algorithm is applicable to both on-policy and off-policy learning. Nevertheless, our primary emphasis lies on off-policy learning. In this section, we introduce our algorithm for the actor and critic. Additionally, we provide a variant of our model that incorporates a natural actor-critic approach.
\subsection{The Critic: Policy Evaluation}
Let V be an approximate value function that can be defined as $V^{\pi}(s_{t})=\mathbb{E}_{a_{t}\in \mathcal{A}}[Q^{\pi}(s_{t},a_{t})]$. The TD residual of V with discount factor $\gamma$ \cite{Sutton2018Reinforcement} is given as $\delta^{V^{\pi}}_{t}=r(s_{t},a_{t}\sim b(.|s_{t}))+\gamma V^{\pi}(s_{t+1}) - V^{\pi}(s_{t})$). $b(.|s)$ is behavior policy probabilities for current state s. Policy gradient uses a value function ($V^{\pi}(s_{t})$) to evaluate a target policy ($\pi$). $\delta^{V^{\pi}}_{t}$ is considered as an estimate of $A^{\pi}_{t}$ of the action $a_{t}$ . i.e., $\delta^{V^{\pi}}_{t}\approx A^{\pi}_{t}$.

\begin{align}
\label{critic7}
\mathbb{E}_{s_{t+1}}[\delta^{V^{\pi}}_{t}] &= \mathbb{E}_{s_{t+1}}[r(s_{t},a_{t}\sim b(.|s_{t}))\nonumber\\
                                           &+\gamma V^{\pi}(s_{t+1}) - V^{\pi}(s_{t})]\\
                                           &= \mathbb{E}_{s_{t+1}}[Q^{\pi}(s_{t},a_{t}) - V^{\pi}(s_{t})]\nonumber \\
                                           &= A^{\pi}(s_{t},a_{t})\nonumber
\end{align}
From the above, it is evident that the agent utilizes the action produced by the behavior policy, rather than the target policy, in our reward approach. The value function is trained using an approximation method to minimize the error in the squared TD residual.
\begin{align}
\label{critic8}
J_{V}(\mathcal{\phi}) &= \mathbb{E}_{s_{t+1}}[\frac{1}{2}(\delta^{V^{\pi}_{\phi}}_{t})^{2}]
\end{align}

\subsection{The Actor: Policy Improvement}
A critic modifies the parameter $\phi$ of the action-value function. The actor adjusts the policy parameter $\theta$ according to the direction suggested by the critic. The actor chooses their course of action, while the critic evaluates the actor's performance and provides feedback on its effectiveness and areas for improvement. The policy gradient can be represented in the following manner:
\begin{align}
J(\theta) &=  \mathbb{E}_{\pi}\Bigg[R(s,a)\Bigg]\nonumber\\
\nabla J(\theta)&= \hat{J}(\theta) = \nabla_{\theta}\mathbb{E}_{\pi}\Bigg[R(s,a)\Bigg]\nonumber\\
\hat{J}(\theta) &= \nabla_{\theta}\sum\limits_{\substack{a\sim\pi}} \pi_{\theta}(a|s)R(s,a) \nonumber\\
\hat{J}(\theta) &= \sum\limits_{\substack{a\sim\pi}} \nabla_{\theta}\pi_{\theta}(a|s)R(s,a) \nonumber\\
\hat{J}(\theta) &= \sum\limits_{\substack{a\sim\pi}} \pi_{\theta}(a|s)\nabla_{\theta}\log\pi_{\theta}(a|s)R(s,a) \nonumber\\
\hat{J}(\theta) &= \sum\limits_{\substack{a\sim\pi}} \pi_{\theta}(a|s)\nabla_{\theta}\log\pi_{\theta}(a|s)R(s,a) \nonumber\\
\hat{J}(\theta) &= \sum\limits_{\substack{a\sim\pi}} \frac{\pi_{\theta}(a|s)}{b(a|s)} b(a|s)\nabla_{\theta}\log\pi_{\theta}(a|s)R(s,a) \nonumber
\shortintertext{From Equation \ref{is2}, Expectation changes to the behavior policy.}\nonumber
\hat{J}(\theta) &= \mathbb{E}_{b}\Bigg[ \frac{\pi_{\theta}(a|s)}{b(a|s)} \nabla_{\theta}\log\pi_{\theta}(a|s)R(s,a)\Bigg] \nonumber
\end{align}
To calculate the policy gradient, we utilize an estimated TD error ($\delta^{V^{\pi}_{\phi}}$) in practical applications. The discounted TD residual ($\delta^{V^{\pi}_{\phi}}$) can be used to construct an off-policy gradient estimator in the subsequent manner.
\begin{align}
\label{actor9}
\hat{J}(\theta) = \frac{1}{N}\sum\limits_{\substack{i=1}}^N \sum\limits_{\substack{t=0}}^\infty \frac{\pi_{\theta}(a_{t}^{i}|s_{t}^{i})}{b(a_{t}^{i}|s_{t}^{i})} \nabla_{\theta}\  log\ \pi_{\theta}(a_{t}^{i}|s_{t}^{i})\delta^{V^{\pi,i}_{\phi}}_{t}
\end{align}
We strive to minimize the instability of off-policy. The disparity between bias and variance (either high bias and high variation or low bias and high variance) typically leads to instability in off-policy scenarios. IS mitigates bias but introduces significant variance. The fluctuation of the IS ratio across different samples is the basis for using IS to average the reward $R(a_{t}|s_{t})\frac{\pi(a_{t}|s_{t})}{b(a_{t}|s_{t})}$, which has a high variance \cite{Hanna2018ImportanceSP,Mahmood2014Weighted,Silver2014DeterministicPG,precup2000eligibility}. Therefore, in order to reduce the impact of large variance (which is directly linked to instability), a smooth version of IS, such as RIS, is necessary. The RIS method exhibits a variance that is constrained within a specific range and a bias that is minimal. Proposition \ref{prop1} has demonstrated the boundedness of RIS, specifically that $\mu_{\beta}\leq\frac{1}{\beta}$. Consequently, the variance of RIS is also bounded. IS is a technique that helps minimize bias. RIS is a modified version of IS that further reduces bias. Therefore, RIS also contributes to bias reduction \cite{hachiya2009adaptive,gu2017interpolated,Mahmood2014Weighted,Sugiyama2016Introduction}. Thus, in order to reduce bias while keeping variance within limits, we employ the off-policy approach. In this approach, we estimate the value of ($J(\theta)$) by utilizing actions chosen from $b(a|s)$ instead of $\pi(a|s)$. We then combine the RIS ratio $\mu_{\beta}$ with $\hat{J}(\theta)$, which we refer to as RIS-off-PAC.
 \begin{align}
\label{actor10}
\hat{J}_{\mu_{\beta}}(\theta) &= \frac{1}{N}\sum\limits_{\substack{i=1}}^N \sum\limits_{\substack{t=0}}^\infty (\frac{e^{\pi(a^{i}_{t}|s^{i}_{t})}}{\beta e^{\pi(a^{i}_{t}|s^{i}_{t})} + (1-\beta)e^{b(a^{i}_{t}|s^{i}_{t})}})\nonumber \\
                              &\quad\quad\quad\quad\quad\quad\nabla_{\theta}\  log\ \pi_{\theta}(a_{t}^{i}|s_{t}^{i})\delta^{V^{\pi,i}_{\phi}}_{t} \nonumber  \\
                              &= \frac{1}{N}\sum\limits_{\substack{i=1}}^N \sum\limits_{\substack{t=0}}^\infty \mu^{i}_{t,\beta} \nabla_{\theta}\  log\ \pi_{\theta}(a_{t}^{i}|s_{t}^{i})\delta^{V^{\pi,i}_{\phi}}_{t}
\end{align}
There are two significant facts that need to be highlighted regarding Equation (\ref{actor10}). Initially, we employ the RIS ($\frac{e^{\pi(a^{i}_{t}|s^{i}_{t})}}{\beta e^{\pi(a^{i}_{t}|s^{i}_{t})} + (1-\beta)e^{b(a^{i}_{t}|s^{i}_{t})}}$) instead of the IS ratio ($\frac{\pi_{\theta}(a_{t}^{i}|s_{t}^{i})}{b(a_{t}^{i}|s_{t}^{i})}$). Secondly, we employ $\mu_{t,\beta}$ in place of $\prod\limits_{\substack{t=0}}^\infty \mu_{t,\beta}$. As a result, it eliminates the need for a product of several unbounded significant weights and instead just requires an approximation of the relative importance weight $\mu_{\beta}$. The bounded RIS is anticipated to exhibit low variance. Here, we introduce two variations of the actor-critic algorithm:
(i) The first method is called relative importance sampling off-policy actor-critic (RIS-off-PAC). (ii) The second method is called relative importance sampling off-policy natural actor-critic (RIS-off-PNAC).
\begin{algorithm}[!htp]
   \caption{The RIS-off-PAC algorithm}
   \label{alg:1}
\begin{algorithmic}
   \STATE {\bfseries Initialize:} policy parameters $\theta$, critic parameters $\phi$, discount factor ($\gamma$), done={\bfseries false}, t=0, $\alpha_{\theta}, \alpha_{\phi}, \beta \in[0,1]$
   \FOR{$i=1$ {\bfseries to} $N$}
   \REPEAT
   \STATE Choose an action $(a_{t}^{i})$, according to $\pi(.|s_{t}^{i}), b(.|s_{t}^{i})$
   \STATE Observe output next state ($\acute{s}^{i}$), reward (r), and done
   \STATE $\mu^{i}_{t,\beta} = \frac{e^{\pi_{\theta}(a^{i}_{t}|s^{i}_{t})}}{\beta e^{\pi_{\theta}(a^{i}_{t}|s^{i}_{t})} + (1-\beta)e^{b(a^{i}_{t}|s^{i}_{t})}}$
   \STATE Update the critic:
   \STATE $\delta^{V^{\pi,i}_{\phi}}_{t}= r(s_{t}^{i},a_{t}^{i}\sim b(.|s_{t}^{i}))+\gamma V^{\pi}_{\phi}(\acute{s}^{i}) - V^{\pi}_{\phi}(s_{t}^{i})$
   \STATE $\nabla_{\phi} J(\phi) \approx  \frac{1}{2}\nabla_{\phi}\|\delta^{V^{\pi,i}_{\phi}}_{t}\|^{2}$
   \STATE $\phi =  \phi + \alpha_{\phi} \nabla_{\phi} J(\phi)$
   \STATE Update the actor:
   \STATE $\nabla_{\theta} J_{\mu_{\beta}}(\theta) \approx  \mu^{i}_{t,\beta} \ \nabla_{\theta}\  log\ \pi_{\theta}(a_{t}^{i}|s_{t}^{i}) \ \delta^{V^{\pi,i}_{\phi}}_{t}$
   \STATE $\theta = \theta + \alpha_{\theta} \nabla_{\theta} J_{\mu_{\beta}}(\theta)$
   \STATE $t \ += 1$
   \STATE $s^{i} = \acute{s}^{i}$
   \UNTIL{$done$ is $false$}
   \ENDFOR
\end{algorithmic}
\end{algorithm}
In algorithms \ref{alg:1} and \ref{alg:2}, $\alpha_{\theta}$ and $\alpha_{\phi}$ represent the learning rates for the actor and critic, respectively. The state s denotes the current state, while the state $\acute{s}$ denotes the subsequent state. The algorithm labeled as \ref{alg:2} is RIS-off-PNAC, which utilizes the natural gradient estimate $\hat{J}_{t}(\theta)=G_{t}^{-1}(\theta)\ \nabla_{\theta}\ log\ \pi_{\theta}(a_{t}|s_{t}) \ \delta^{V^{\pi}_{\phi}}_{t}$. The natural gradient $G^{-1}_{t}(\theta)$ is discussed in more detail in the following references: \cite{Bhatnagar2009NaturalAA,Konda2003OnActorCriticA,Peters2005NaturalA,Silver2014DeterministicPG}. The sole distinction between RIS-off-PAC and RIS-off-PNAC lies in the substitution of the regular gradient estimate with the natural gradient estimate in RIS-off-PNAC. The RIS-off-PNAC algorithm \ref{alg:2} employs Equation 26 from the reference \cite{Bhatnagar2009NaturalAA} to calculate the natural gradient. While the natural actor-critic (NAC) methods proposed by \cite{Bhatnagar2009NaturalAA} are on-policy, our algorithm operates off-policy. In the field of RL, our objective is to maximize the rewards. Therefore, the problem we are addressing here is an optimization problem focused on maximizing rather than minimizing. In the original problem, we aim to maximize the reward by minimizing a negative loss function, which is equivalent to finding the largest reward.
\begin{algorithm}[!htp]
   \caption{The RIS-off-PNAC algorithm}
   \label{alg:2}
\begin{algorithmic}
   \STATE {\bfseries Initialize:} policy parameters $\theta$, critic parameters $\phi$, discount factor ($\gamma$), done={\bfseries false}, t=0, $\alpha_{\theta}, \alpha_{\phi}, \beta \in[0,1], G_{0}=\mathbb{I}$
   \FOR{$i=1$ {\bfseries to} $N$}
   \REPEAT
   \STATE Choose an action $(a_{t}^{i})$, according to $\pi(.|s_{t}^{i}), b(.|s_{t}^{i})$
   \STATE Observe output next state ($\acute{s}^{i}$), reward (r), and done
   \STATE $\mu^{i}_{t,\beta} = \frac{e^{\pi_{\theta}(a^{i}_{t}|s^{i}_{t})}}{\beta e^{\pi_{\theta}(a^{i}_{t}|s^{i}_{t})} + (1-\beta)e^{b(a^{i}_{t}|s^{i}_{t})}}$
   \STATE Update the critic:
   \STATE $\delta^{V^{\pi,i}_{\phi}}_{t}= r(s_{t}^{i},a_{t}^{i}\sim b(.|s_{t}^{i}))+\gamma V^{\pi}_{\phi}(\acute{s}^{i}) - V^{\pi}_{\phi}(s_{t}^{i})$
   \STATE $\nabla_{\phi} J(\phi) \approx  \frac{1}{2}\nabla_{\phi}\|\delta^{V^{\pi,i}_{\phi}}_{t}\|^{2}$
   \STATE $\phi =  \phi + \alpha_{\phi} \nabla_{\phi} J(\phi)$
   \STATE Update the actor:
   \STATE $G^{-1}_{t}(\theta) = \frac{1}{1-\alpha_{\theta,t}} \Bigg[G^{-1}_{t-1}(\theta) -  \alpha_{\theta,t} \frac{(G^{-1}_{t-1}(\theta) \nabla_{\theta}\log \pi_{\theta}(a_{t}^{i}|s_{t}^{i})) \ (G^{-1}_{t-1}(\theta) \nabla_{\theta}\log \pi_{\theta}(a_{t}^{i}|s_{t}^{i}))^{T}}{1-\alpha_{\theta,t}+\alpha_{\theta,t} (\nabla_{\theta}\log \pi_{\theta}(a_{t}^{i}|s_{t}^{i}))^{T} G^{-1}_{t-1}(\theta) \nabla_{\theta} \log \pi_{\theta}(a_{t}^{i}|s_{t}^{i})}\Bigg]$
   \STATE $\nabla_{\theta} J_{\mu_{\beta}}(\theta) \approx  \mu^{i}_{t,\beta} \ \nabla_{\theta}\  log\ \pi_{\theta}(a_{t}^{i}|s_{t}^{i}) \ G^{-1}_{t}(\theta)\ \delta^{V^{\pi,i}_{\phi}}_{t}$
   \STATE $\theta = \theta + \alpha_{\theta} \nabla_{\theta} J_{\mu_{\beta}}(\theta)$
   \STATE $t \ += 1$
   \STATE $s^{i} = \acute{s}^{i}$
   \UNTIL{$done$ is $false$}
   \ENDFOR
\end{algorithmic}
\end{algorithm}
\begin{lemma}
\label{lemma1}
The RIS estimator ($\hat{\mu}_{\beta}$) becomes the ordinary IS estimator ($\hat{\mu}_{b}$) if $\beta = 0$.\\
The proof is presented in appendix \ref{proofs}.
\end{lemma}

\begin{prop}
\label{prop2}
If $\beta = 0$, the RIS off-policy gradient estimator becomes the ordinary IS off-policy gradient estimator.\\
The proof is presented in appendix \ref{proofs}.
\end{prop}

\begin{lemma}
\label{lemma2}
The RIS estimator produces uniform weight $\hat{\mu}_{\beta}=\frac{1}{1-\gamma}$ if $\beta = 1$.\\
The proof is presented in appendix \ref{proofs}.
\end{lemma}

\begin{lemma}
\label{lemma3}
The RIS produces uniform weight 1 if $\beta = 1$.\\
The proof is presented in appendix \ref{proofs}.
\end{lemma}

\begin{prop}
\label{prop3}
If $\beta = 1$, the RIS off-policy gradient estimator becomes the ordinary on-policy gradient estimator.\\
The proof is presented in appendix \ref{proofs}.
\end{prop}
\setcounter{theorem}{0} 
\begin{theorem}
\label{theorem2}
As $\beta$ increases from 0 to 1, the variance of the RIS estimator, $V_{\beta}(\hat{\mu}_{\beta})$, decreases, reaching zero when $\beta = 1$. The relationship is given by:\\\\
$V_{\beta}(\hat{\mu}_{\beta}) = \frac{2\gamma(1-\gamma)(1-\beta)}{[\beta\pi(A|S) + (1-\beta)b(A|S)]^{2}}$\\\\
The proof is presented in appendix \ref{proofs}.
\end{theorem}

\begin{theorem}
\label{theorem3}
If $\beta = 0$, then the variance of RIS estimator ($Var_{b}(\hat{\mu}_{\beta})$) is $\mathbb{E}_{b}[\hat{\mu}_{b}^{2}]$. This Theorem captures the variance of the RIS estimator for any general value of \textit{$\beta$ from 0 to 1}.\\
The proof is presented in appendix \ref{proofs}.
\end{theorem}

\begin{remark}
\label{remark1}
If $\beta=0$, Lemma \ref{lemma1} shows that RIS estimator is equal to standard IS estimator. Theorem \ref{theorem3} also shows that variance of RIS estimator is also equal to standard IS estimator when $\beta=0$. Therefore, we conclude that if the expectation of RIS and standard IS are equal, then their variances are also equal.
\end{remark}

\begin{theorem}
\label{theorem4}
If $\beta = 1$, Then, the variance of RIS estimator ($\hat{\mu}_{\beta})$) is $\frac{-2\gamma}{(1-\gamma^{2})(1-\gamma)}$.\\
The proof is presented in appendix \ref{proofs}.
\end{theorem}

\begin{theorem}
\label{theorem5}
If $\beta = 1$, Then, the variance of RIS is zero.\\
The proof is presented in appendix \ref{proofs}.
\end{theorem}

\begin{remark}
\label{remark2}
$\beta[0,1]$ controls the smoothness. The RIS ($\mu_{\beta}$) becomes the ordinary IS ($\frac{\pi(a|s)}{b(a|s)}$) if $\beta = 0$. RIS becomes smoother if $\beta$ is increased, and it produces uniform weight $\mu_{\beta}=1$ if $\beta = 1$. It is proved by Lemma \ref{lemma1} and \ref{lemma3}. Smoothness is directly proportional to the value of $\beta$. Variance decreases when smoothness rises. Therefore, Smoothness is directly proportional to the stability of off-policy. Thus, $\beta$ controls the stability of off-policy, as $\beta$ increases off-policy becomes more stable.
\end{remark}

\begin{remark}
\label{remark3}
The RIS estimator $\hat{\mu}_{\beta}$  is a reliable and unbiased estimate of $\pi$. The bounded variance of $\hat{\mu}_{\beta}$ is due to the boundedness of RIS, as stated in proposition \ref{prop1}. The conventional IS estimator is unbiased, but it is plagued by significant variance due to the multiplication of numerous unbounded importance weights \cite{Wang2016SampleEA,hachiya2009adaptive}. However, RIS exhibits low variance due to the absence of a multiplication involving numerous unbounded weights.
\end{remark}

\subsection{RIS-Off-Policy Actor-critic Architecture}
\begin{figure}[!htbp]
\centering
\subfigure[The RIS-Off-PAC Architecture.]{
\includegraphics[width=0.5\columnwidth]{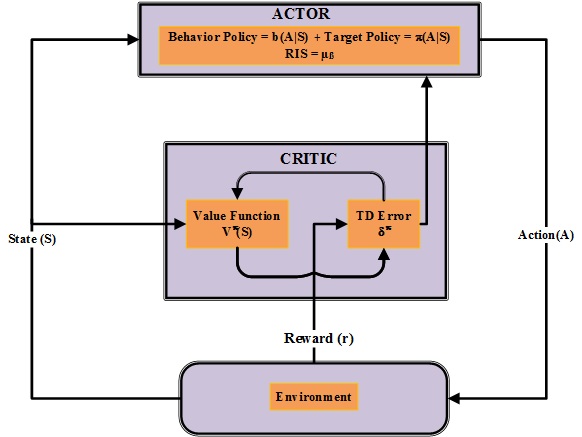}
\label{ac-arch}
}%
\subfigure[The RIS-Off-PAC Neural Network Architecture.]{
\includegraphics[width=0.5\columnwidth]{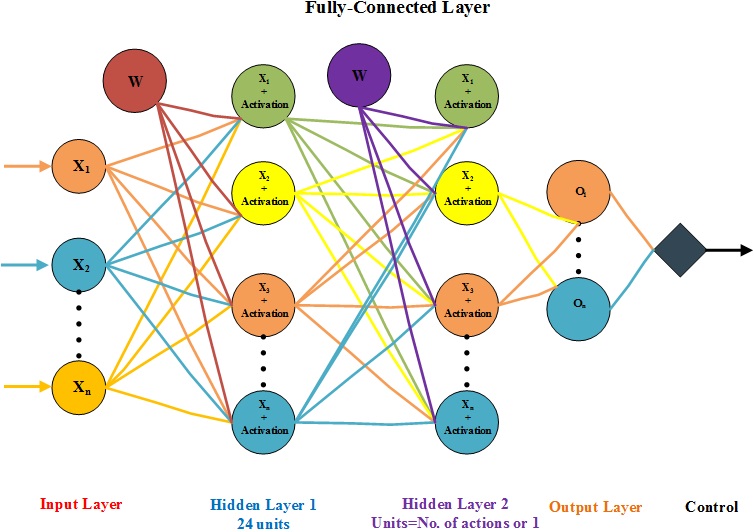}
\label{nn-arch}
}
\caption{Illustration of RIS-Off-PAC Architectures.}
\label{arch}
\end{figure}

Fig.\ref{ac-arch} depicts the architecture of RIS-off-PAC. The distinction between RIS-off-PAC and the conventional actor-critic architecture \cite{Sutton1999PolicyGM,Sutton2018Reinforcement} lies in the incorporation of a behavior policy based on RIS in our approach. Instead of using $\pi(A|S)$, we utilize the action created by $b(A|S)$ in the reward function. We calculate the RIS by incorporating both the $\pi(A|S)$ and $b(A|S)$ policies into an actor. Consequently, we provide samples from $b(A|S)$ to the actor, as depicted in Fig.\ref{ac-arch}. The TD error and other factors are identical to those of a conventional actor-critic approach.\par

Fig.\ref{nn-arch} shows the RIS-off-PAC neural network (NN) architecture. We use control RL tasks: CartPole-v0, MountainCar-v0, Pendulum-v0 and Humanoid-v2 for our experiment. We apply our RIS-off-PAC-NN on all of these tasks. Details of our NN as follows:  In our architecture, we have a target network (Actor), value network (Critic) and off-policy network (behavior policy). Each of them implemented as a fully connected layer using TensorFlow as shown in Fig.\ref{nn-arch}. Each NN contains inputs layer, 2 hidden layers: hidden layer 1 and hidden layer 2, and an output layer. Hidden layer 1 has 24 neurons (units) for all three Network for all RL task. Hidden layer 2 has a single neuron in the value network for all RL task. A number of neurons in hidden layer 2 for target network and off-policy network are equal to a number of actions available in given RL task.  Hidden layer 1 employs RELU activation function in target and value network while CRELU activation function used in the off-policy network. Hidden layer 2 utilizes SOFTMAX activation function in target and off-policy network whereas it uses no activation function in the value network. Weight W is generated using the "he\_uniform" function of TensorFlow for all NN and tasks. We availed AdamOptimizer for learning neural network parameters for all RL tasks. $\beta$ is generated uniform random values between 0 and 1. We set numpy random seed, TensorFlow random seed and OpenAI Gym environment seed to 1 to reproduce results.
\section{Empirical results and analysis}
\label{Empirical}
\subsection{Variance versus Beta}
\label{VB}
\begin{figure}[!htbp]
\includegraphics[width=1.0\columnwidth]{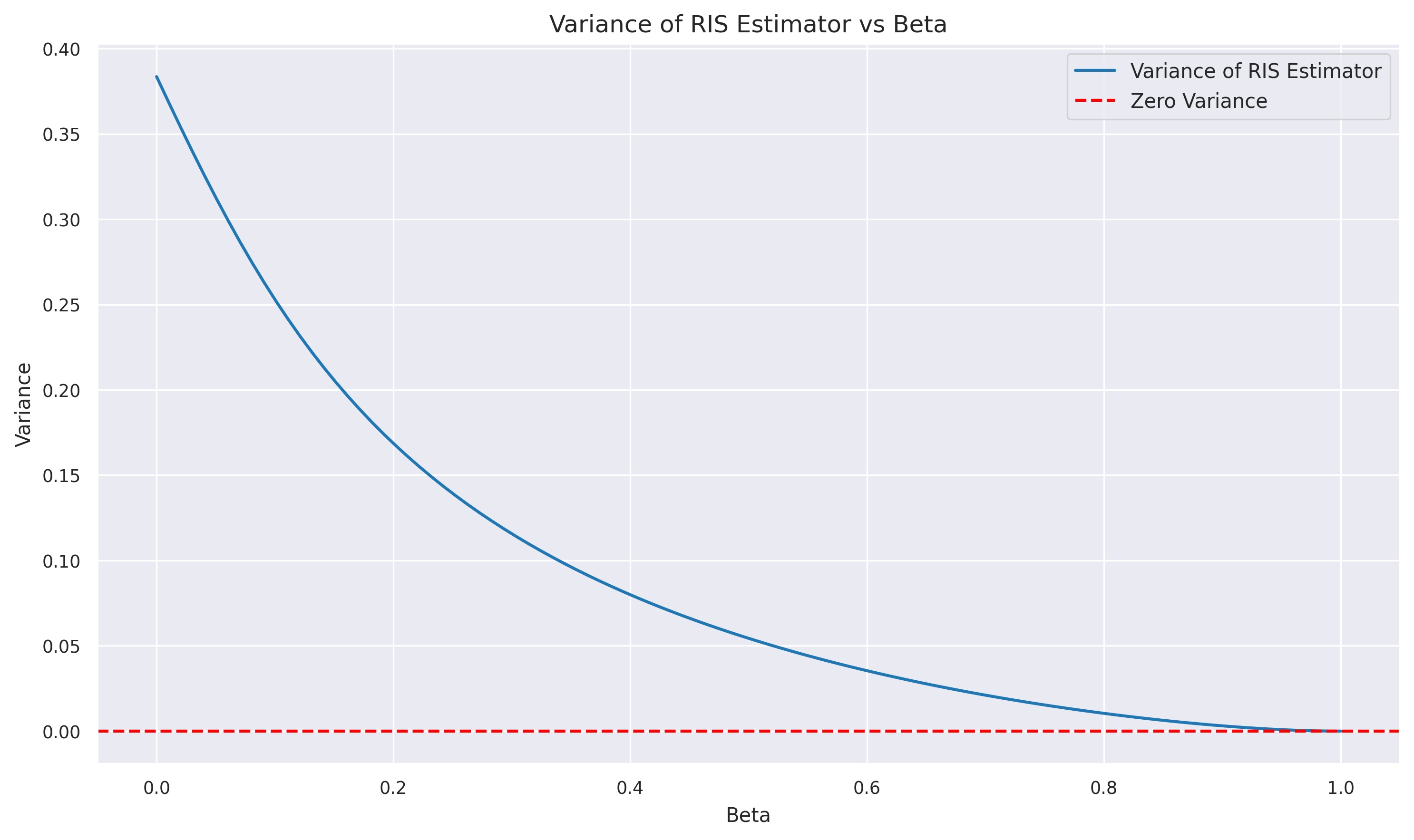}
\caption{Variance of the RIS estimator in relation to beta values}\label{variance}
\end{figure}

We synthetically simulate Theorem \ref{theorem2} to illustrate that when the beta values increase, the variance diminishes. Fig.\ref{variance} distinctly demonstrates these findings. The subsequent parameters were employed to execute the experiment: Discount factor, $\gamma = 0.99$; Number of actions = 5; Actions for both behaviour and target policies were created randomly; Beta values were randomly generated within the range of 0 to 1.
We conducted 1000 simulations for each beta value, estimated the variance for each, and subsequently determined the mean variance for each beta value. The RIS estimator utilised in this experiment derives from the formula presented in Theorem \ref{theorem2}.
\subsection{Empirical Comparison with Q-Estimators}
\label{QE}
\begin{figure}[H]
\centering
\subfigure[Uniform Behavior]{
\includegraphics[width=0.5\columnwidth]{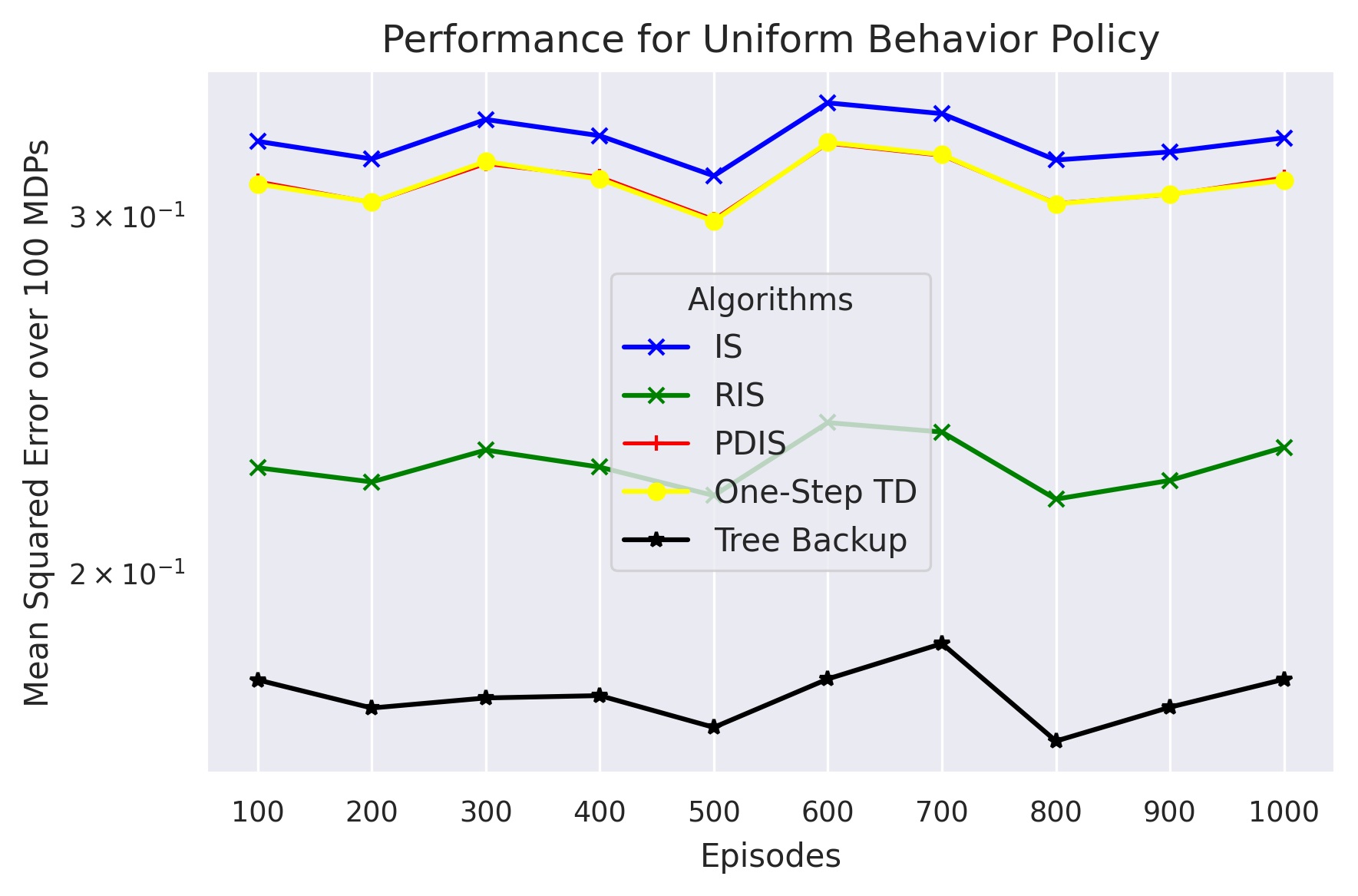}
\label{UB}
}%
\subfigure[Different Behavior]{
\includegraphics[width=0.5\columnwidth]{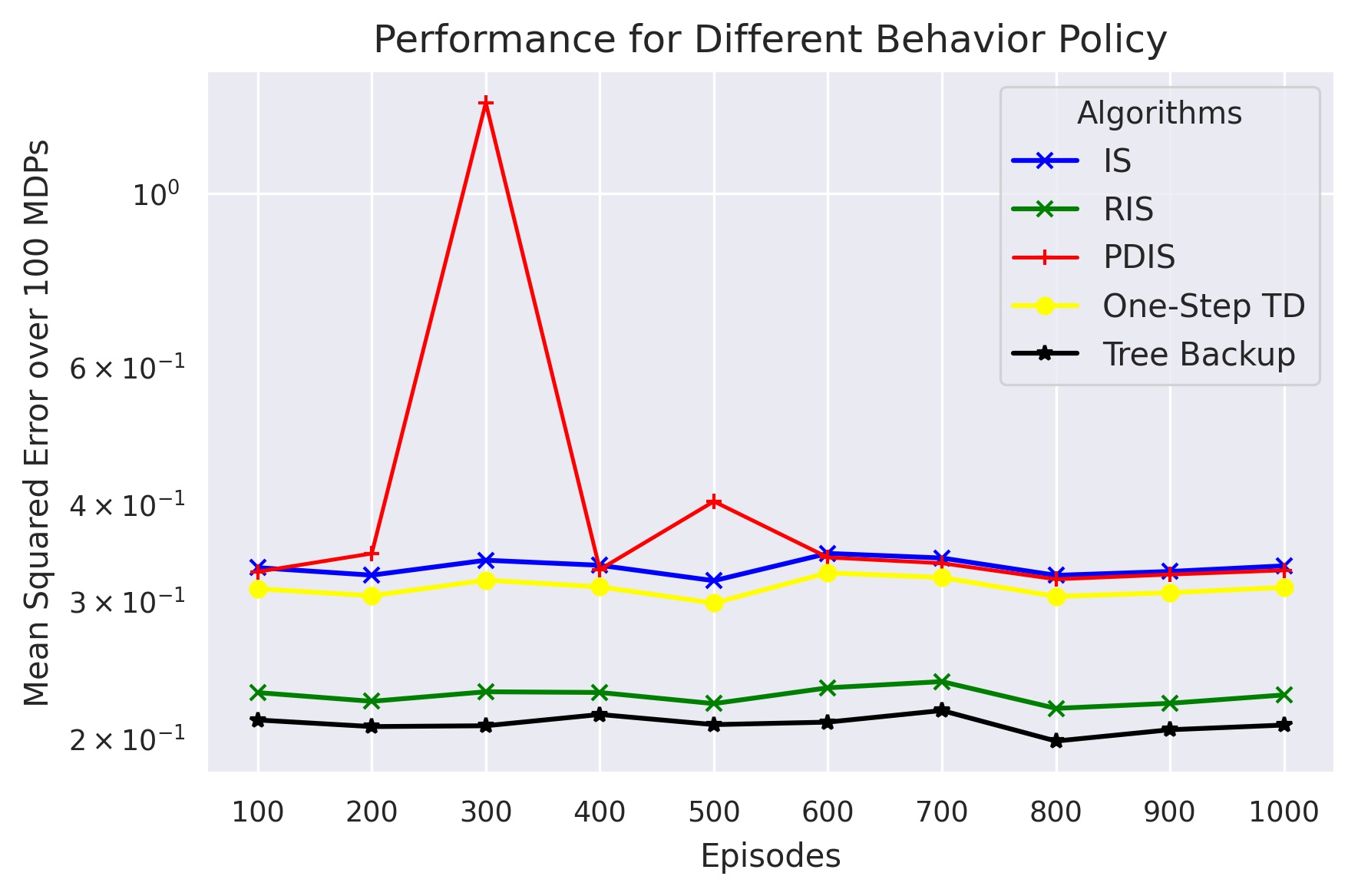}
\label{DB}
}%
\caption{Aggregate performance of all algorithms. The behaviour policy on the left choose between the two actions with an equal probability of 50\%. The behaviour policy selected actions with an 80\%-20\% probability distribution, directly contrasting the target policy's choices.}
\end{figure}

In this experiment, we employed the identical arrangement outlined in the cited study\cite{precup2000eligibility}. We evaluated the estimators using a set of 100 randomly generated MDPs, each comprising 100 non-terminal states, one terminal state, with a gamma of 1.0, and alpha of 0.001. In any non-terminal state, two actions were available, each leading to four randomly chosen subsequent states with assigned random probability. The objective policy was to choose the initial action with an 80\% likelihood and the subsequent action with a 20\% likelihood. The immediate rewards were selected evenly at random from the interval [0, 1].
Two distinct behaviour policies were employed: in the uniform behaviour scenario, both actions were executed with equal probability of 50\%, whereas in the different behaviour policy, the first action was chosen with a 20\% probability and the second with an 80\% probability, leading to a policy that markedly diverged from the target policy. Performance metrics are displayed for a maximum of 1,000 episodes for both the uniform and distinct behaviour policy in Fig.\ref{UB} and Fig.\ref{DB}, respectively. We computed a moving average with a window size of 100 to refine the results.

We acquired empirical data utilising the explicit estimators: $Q^{RIS}$, $Q^{IS}$, $Q^{PDIS}$, the one-step TD approach, and a tree backup method. Our investigation demonstrates that $Q^{RIS}$ exhibits a lower mean squared error (MSE) than all other algorithms, with the exception of the tree backup technique, in both uniform and different behaviour policy scenarios, as depicted in Fig.\ref{UB} and Fig.\ref{DB}. This signifies that $Q^{IS}$, $Q^{PDIS}$, and the one-step TD technique exhibit greater variance.

\subsection{Gap Factor Between Target and Behavior Policy}
\begin{figure}[!htb]
\centering
\subfigure[Rewards]{
\includegraphics[width=0.5\columnwidth]{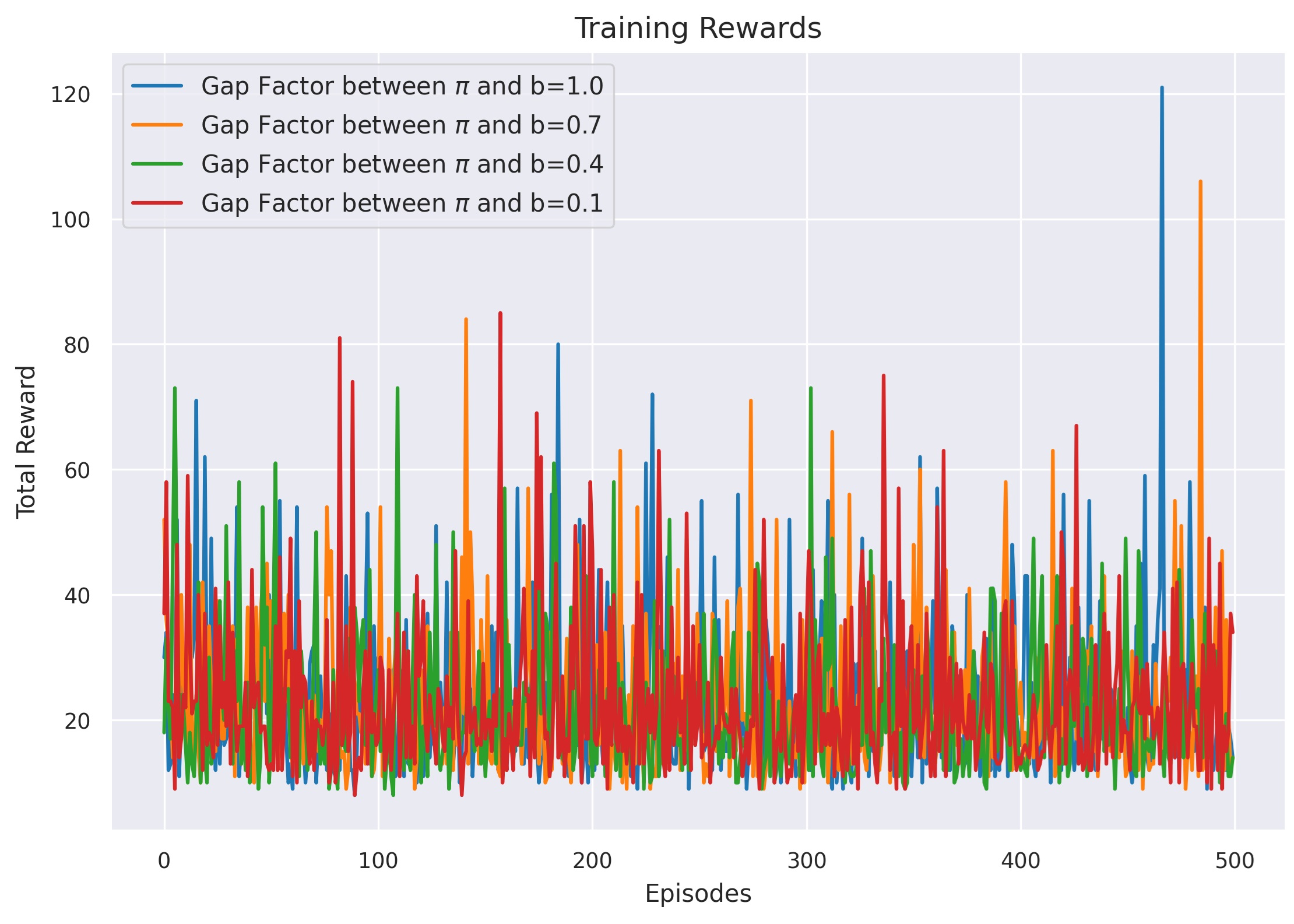}
\label{GReward}
}%
\subfigure[Variances]{
\includegraphics[width=0.5\columnwidth]{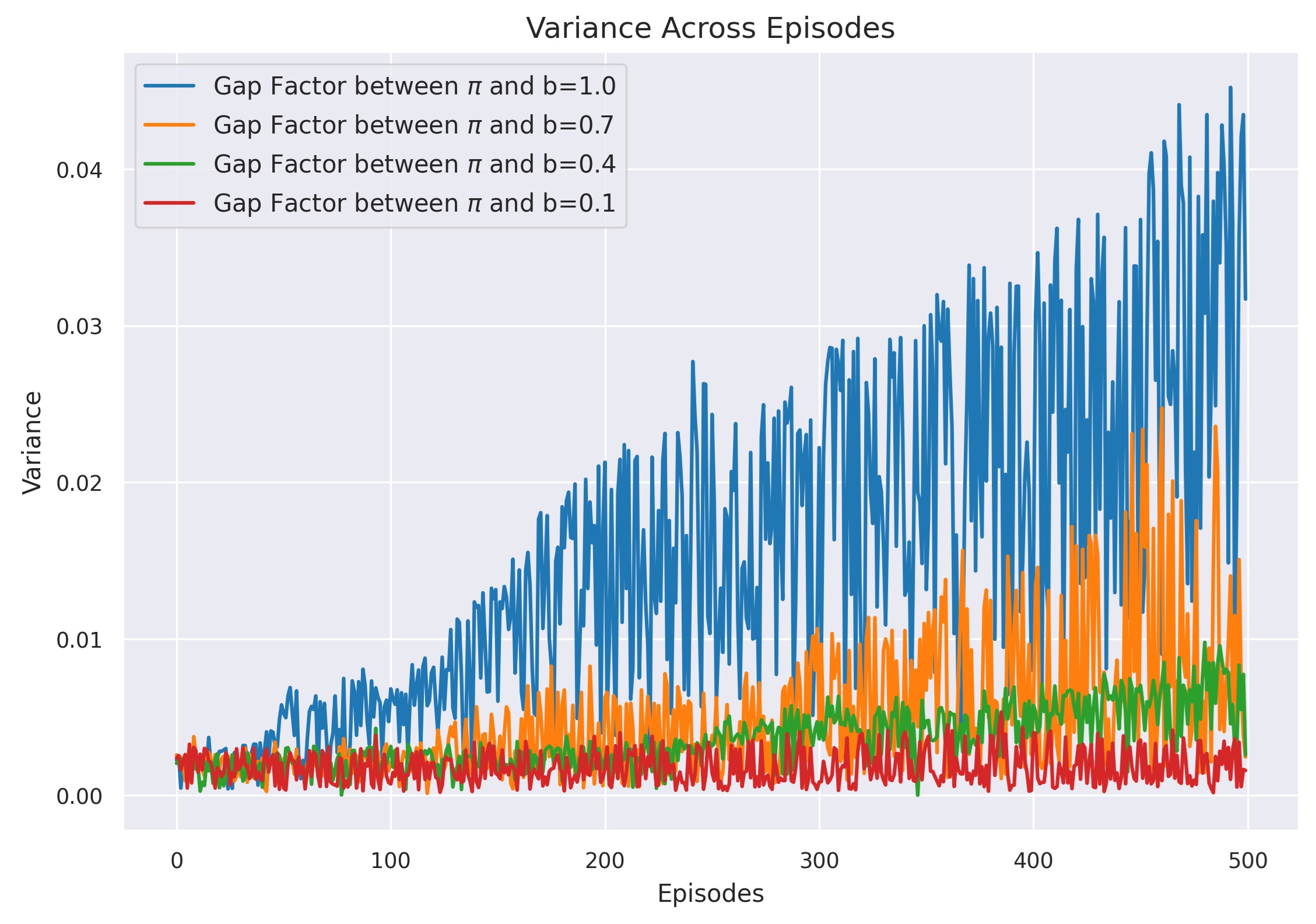}
\label{GVar}
}%
\caption{Off-Policy learning with large gap between target and behavior Policy.}
\end{figure}

Subsequently, we execute RIS off-policy learning on the CartPole-v1 environment from OpenAI Gym. We executed 500 simulations for each value of the gap factor. All remaining configurations are detailed in appendix \ref{poleapp}. This experiment investigates the impact of enlarging the disparity between the target and behaviour policies. The simulation findings indicate that when the gap widens, variance escalates, as depicted in Fig.\ref{GVar}, with the associated rewards represented in Fig.\ref{GReward}. A significant variance in rewards indicates that a greater disparity between behaviour and target policies results in more variance in total rewards across episodes, underscoring instability in off-policy learning. The instability in off-policy learning occurs because of the significant variance in importance sampling weights when there is a substantial disparity between the behaviour and target policies.
\label{GF}

\subsection{Stable Learning versus Variance}
\label{SLV}
we execute RIS-off-PAC and RIS-off-PNAC off-policy learning on the Pendulum-v1 environment from OpenAI Gym. We executed 500 simulations for each value of the beta. All remaining configurations are detailed in appendix \ref{pendulumapp}. In both Fig.\ref{PACRV} and \ref{PNACRV}, total rewards shown in left side while corresponding variance shown in right side. Both Figures show that total reward (i.e. learning) is stable while over all variance is decreasing when beta is increasing, especially when beta = 0.8 and 0.1.
\begin{figure}[!htbp]
\includegraphics[width=1.0\columnwidth]{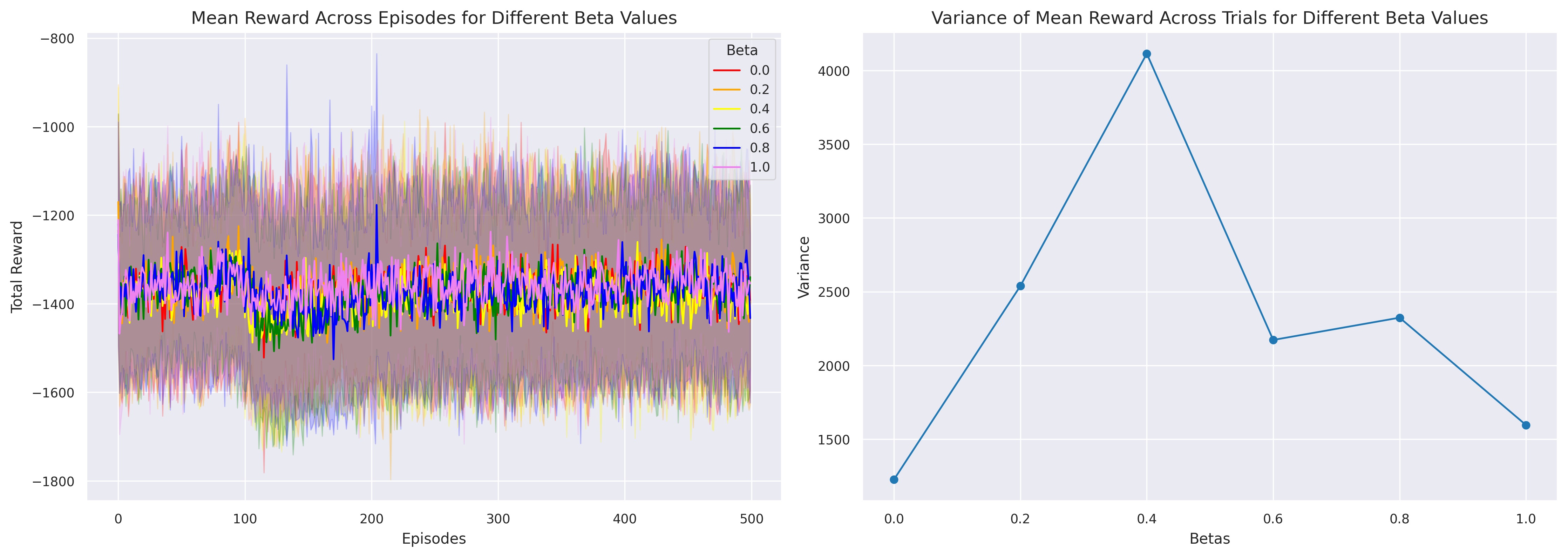}
\caption{RIS-off-PAC exhibit reward outcomes and their associated variance in the Pendulum environment.}\label{PACRV}
\end{figure}

\begin{figure}[!htbp]
\includegraphics[width=1.0\columnwidth]{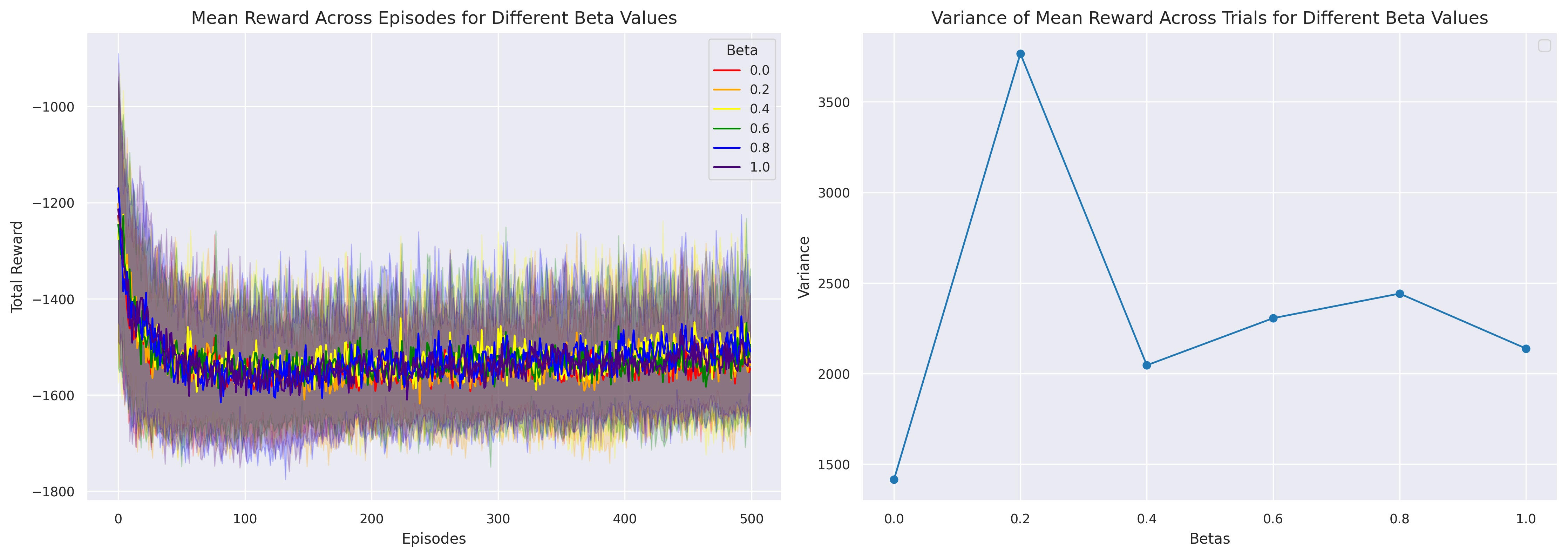}
\caption{RIS-off-PNAC exhibit reward outcomes and their associated variance in the Pendulum environment.}\label{PNACRV}
\end{figure}

\section{Experimental Setup}
\label{experiment}
\begin{figure}[!htbp]
\centering
\subfigure[CartPole v0]{
\includegraphics[width=0.25\columnwidth]{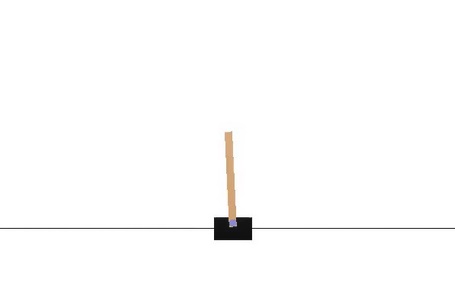}
\label{envpole}
}%
\subfigure[MountainCar v0]{
\includegraphics[width=0.25\columnwidth]{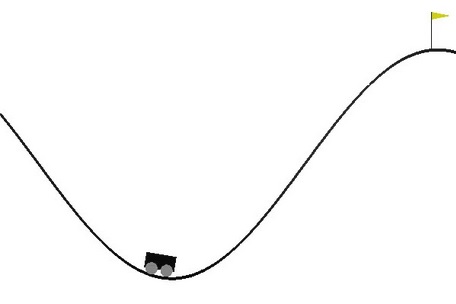}
\label{envcar}
}%
\subfigure[Pendulum v0]{
\includegraphics[width=0.25\columnwidth]{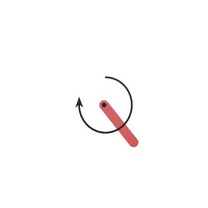}
\label{envpendulum}
}
\subfigure[Humanoid v2]{
\includegraphics[width=0.15\columnwidth]{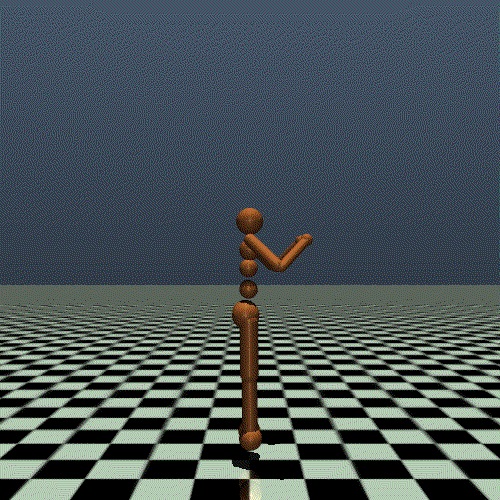}
\label{envlunar}
}%
\caption{We used OpenAI Gym control for all experiments. The experiments were conducted in the following order: CartPole, MountainCar, Pendulum, and Humanoid-v2. Detailed descriptions of each environment are provided in the appendices \ref{poleapp}, \ref{carapp}, \ref{pendulumapp}, and \ref{humanoidapp}.}
\label{allenv}
\end{figure}

We conducted experiments on OpenAI Gym control tasks. The depicted environments are illustrated in Fig.\ref{allenv}. The studies are conducted on a solitary PC equipped with 16 GB of memory, an Intel Core i7-2600 CPU, and GPU. The operating system we utilized was 64-bit Ubuntu 18.04.1 LTS. For programming, we employed Python 3.6.4 and the TensorFlow 1.7 library. Additionally, we made use of the OpenAI Gym toolkit, as referenced in \cite{Brockman2016OpenAIG}. All experiments utilised five random seeds, and the average outcomes are presented.
\subsection{Experimental Results}
We evaluated RIS-off-PAC and RIS-off-PNAC algorithms on four OpenAI Gym's environments: CartPole-v0, MountainCar-v0, Pendulum-v0 and Humanoid-v2. We compared the proposed methods with the following algorithms: asynchronous advantage actor-critic (A3C) \cite{Mnih2016Asynchronous}, proximal policy optimization (PPO) \cite{schulman2017proximal}, policy gradient soft-max (PG) \cite[Chapter 13]{Sutton2018Reinforcement} and soft actor-critic (SAC) \cite{Haarnoja2018SoftAO}.\par

\begin{figure}[!htb]
\centering
\subfigure[MountainCar]{
\includegraphics[width=0.5\columnwidth]{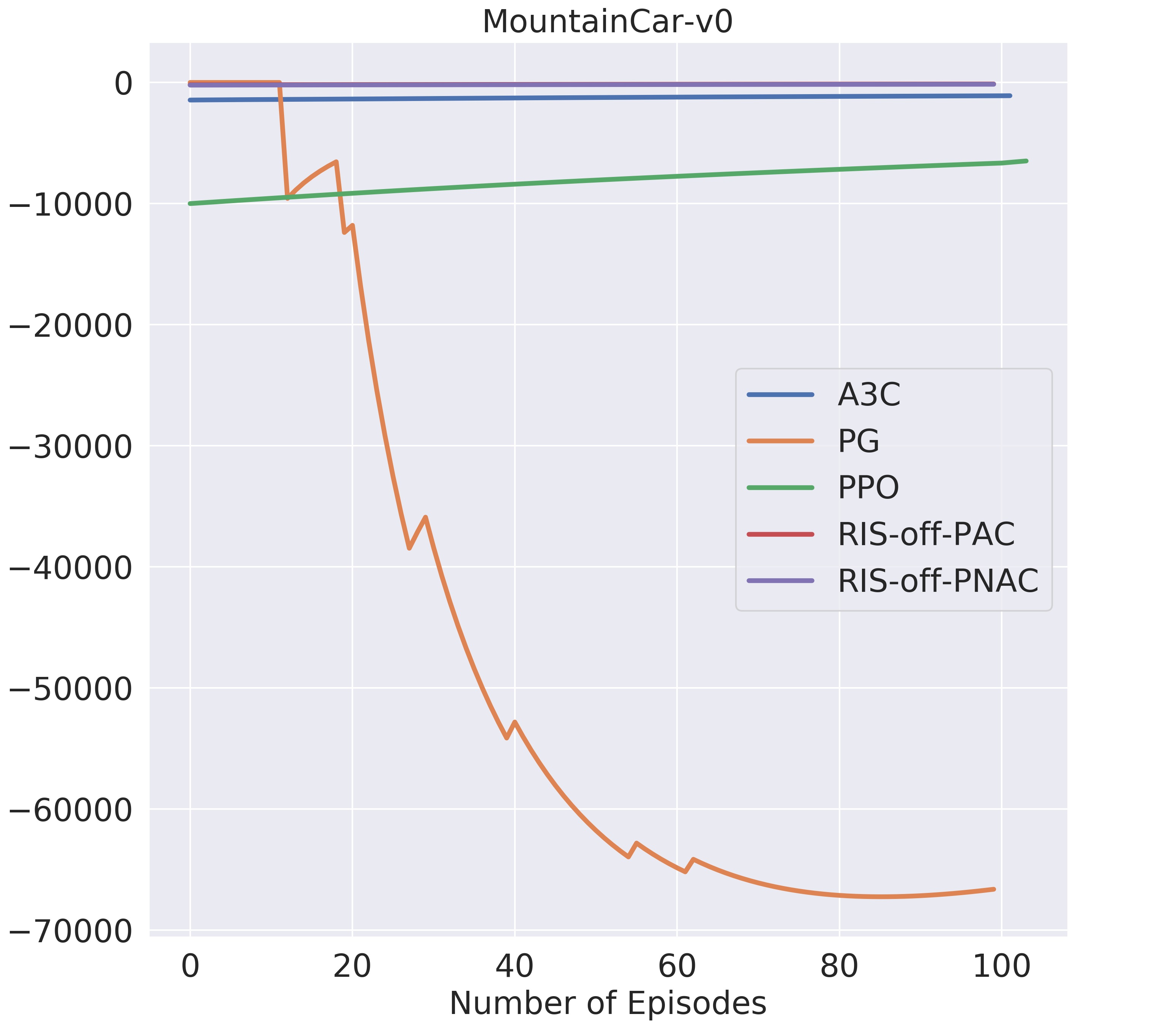}
\label{carfigmain}
}%
\subfigure[Pendulum]{
\includegraphics[width=0.5\columnwidth]{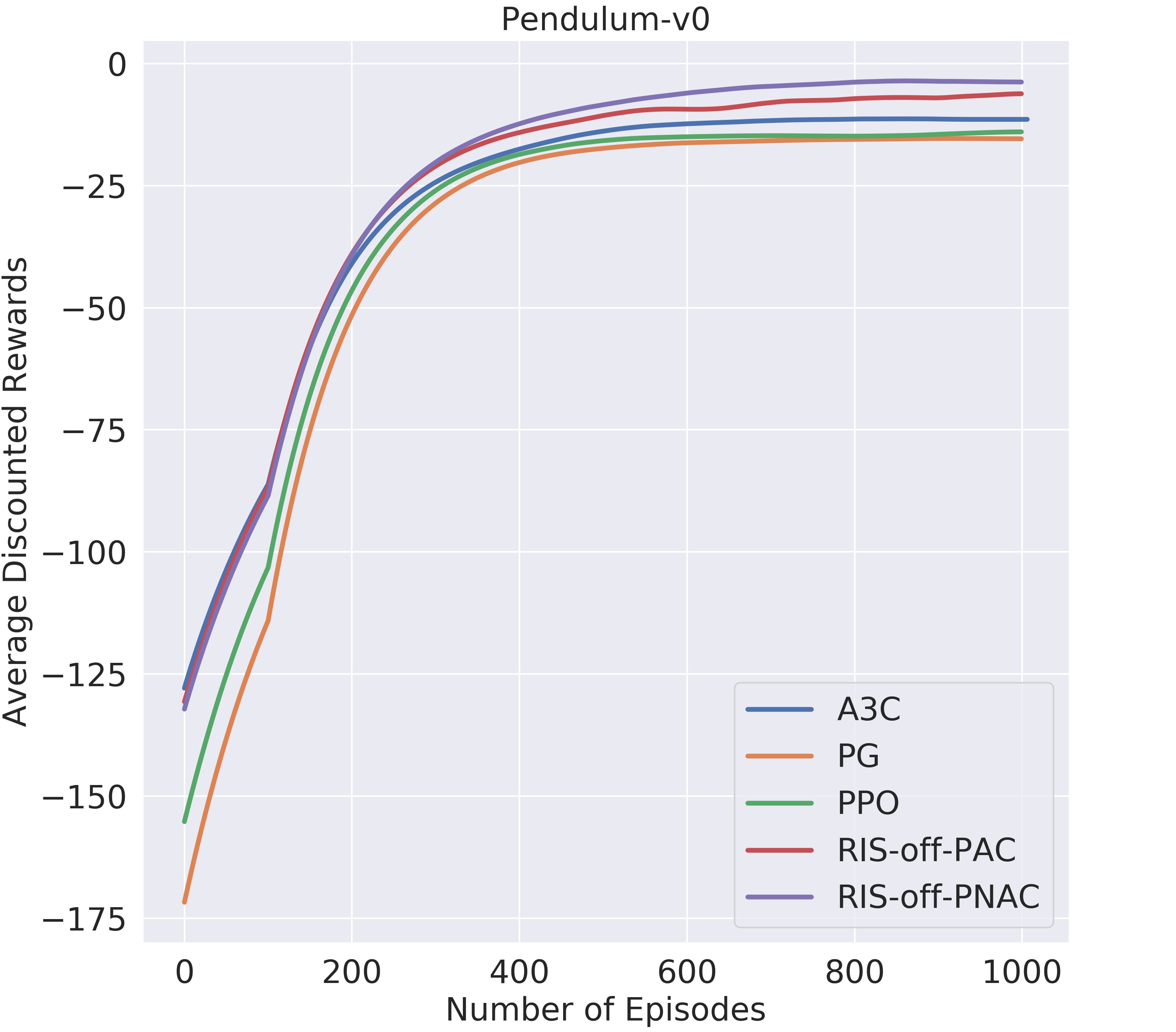}
\label{pendulumfigmain}
}
\caption{(a) Training summary of all algorithms of MountainCar. (b) Training summary of all algorithms of Pendulum. The x-axis shows the total number of training episodes. The y-axis denotes the averaged rewards for MountainCar and Pendulum over 100 and 1000 episodes respectively.}
\end{figure}

\begin{figure}[!htb]
\centering
\subfigure[RIS-off-PAC Algorithm]{
\includegraphics[width=0.5\columnwidth]{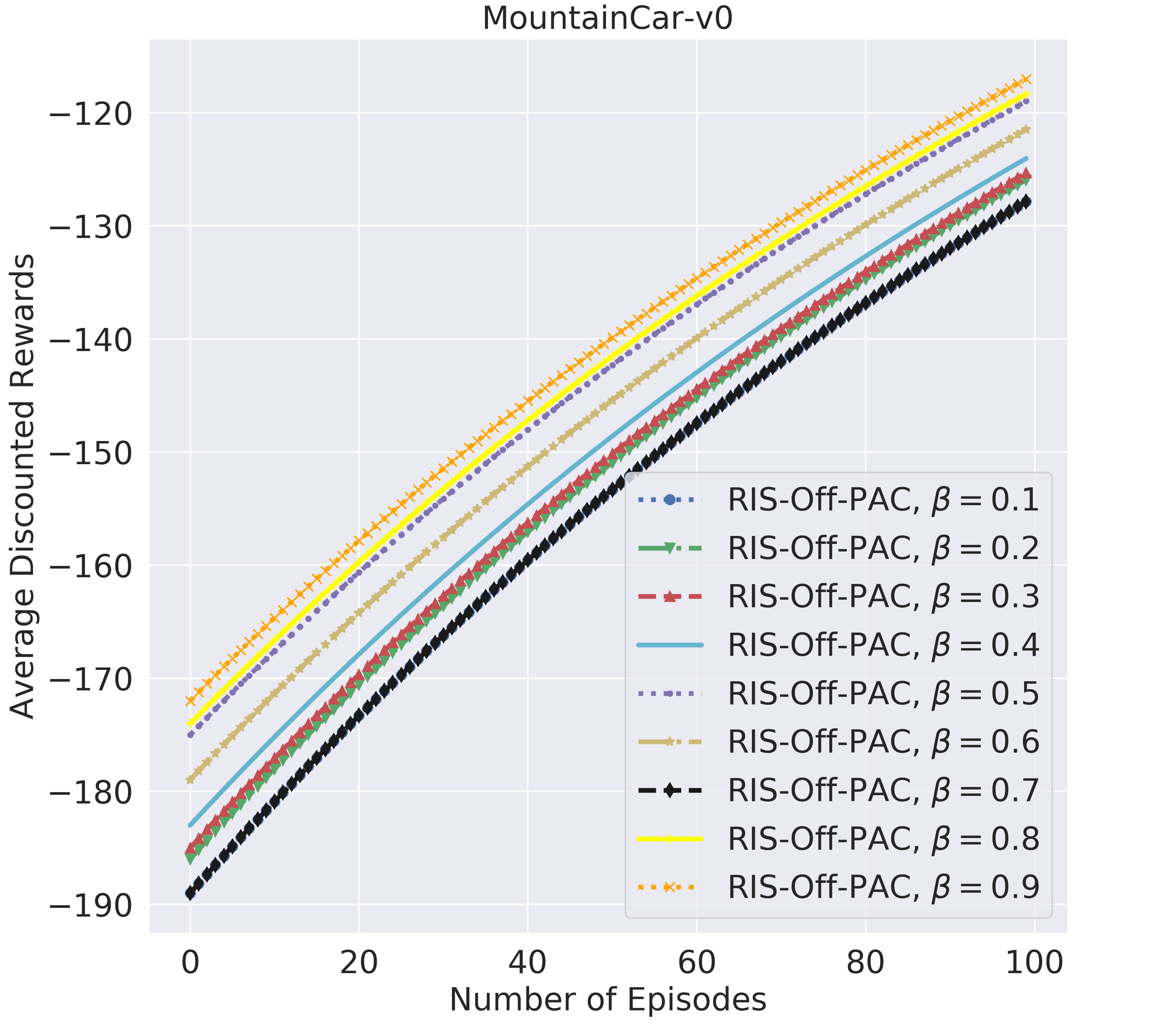}
\label{carfig-ris-off-pac}
}%
\subfigure[RIS-off-PNAC Algorithm]{
\includegraphics[width=0.5\columnwidth]{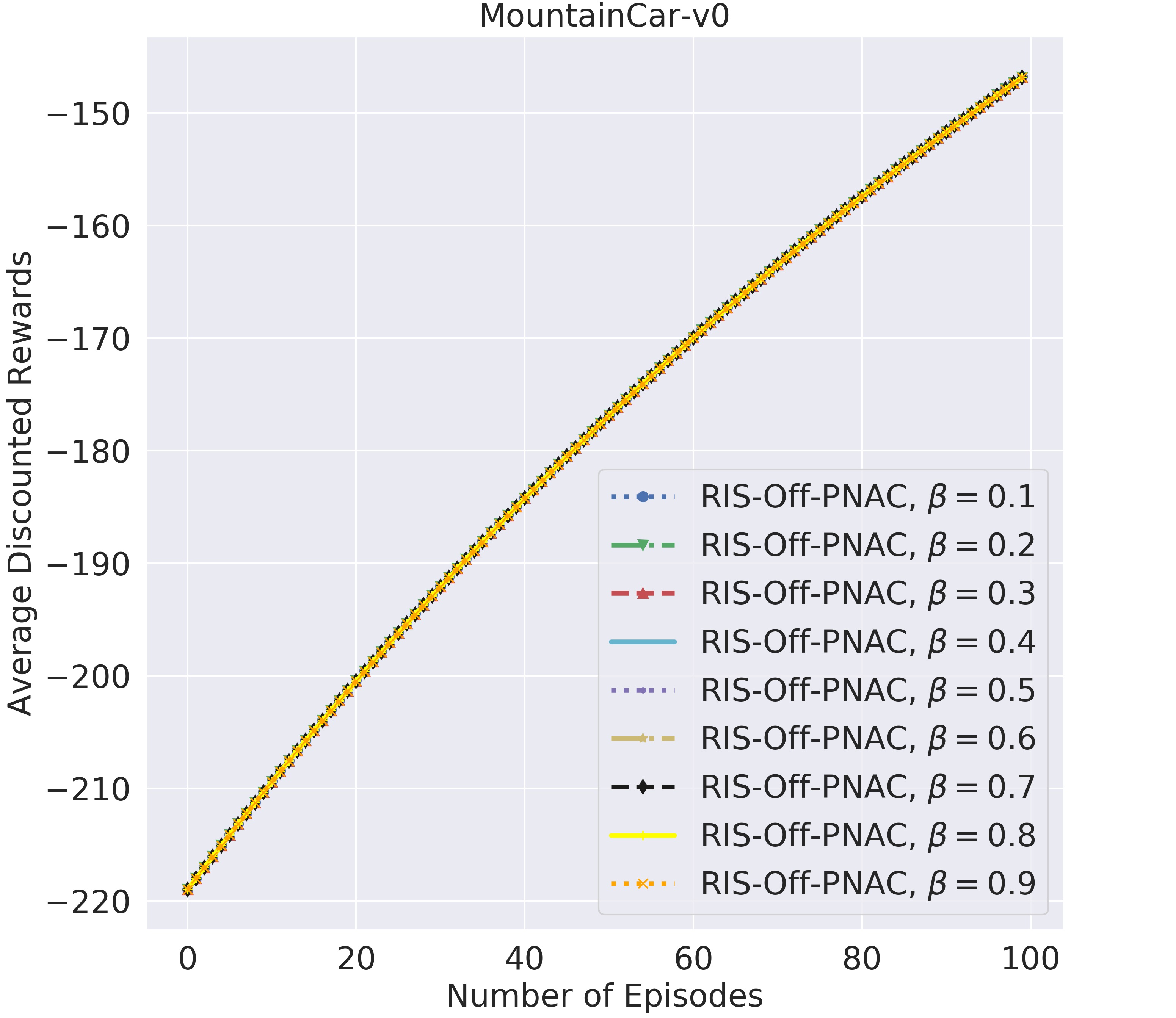}
\label{carfig-ris-off-pnac}
}

\caption{(a), (b) Training summary of RIS-off-PAC and RIS-off-PNAC respectively for different value of $\beta\in[0,1]$. The x-axis shows the total number of training episodes. The y-axis shows the averaged rewards over 100 episodes.}
\end{figure}

The goal of MountainCar-v0 is to drive up on the right and reach the top of the mountain in the fewest number of attempts and steps possible. Our algorithms are limited to a maximum of 100 episodes. Fig.\ref{carfigmain} displays the mean reward obtained by all methods. Fig.\ref{carfigmain} demonstrates that both the RIS-off-PAC and RIS-off-PNAC algorithms beat all other methods. The outcomes of RIS-off-PAC and RIS-off-PNAC exhibit a high degree of similarity. The outcomes of the RIS-off-PAC and RIS-off-PNAC algorithms, utilizing various values of $\beta$, are displayed in Fig.\ref{carfig-ris-off-pac} and Fig.\ref{carfig-ris-off-pnac} correspondingly. In general, the results of RIS-off-PNAC are the most consistent for all values of $\beta$, as depicted in Fig.\ref{carfig-ris-off-pnac}. Fig.\ref{carfig-ris-off-pac} demonstrates the consistent stability of the RIS-off-PAC results across all $\beta$ levels.\par

\begin{figure}[!htb]
\centering
\subfigure[RIS-off-PAC Algorithm]{
\includegraphics[width=0.5\columnwidth]{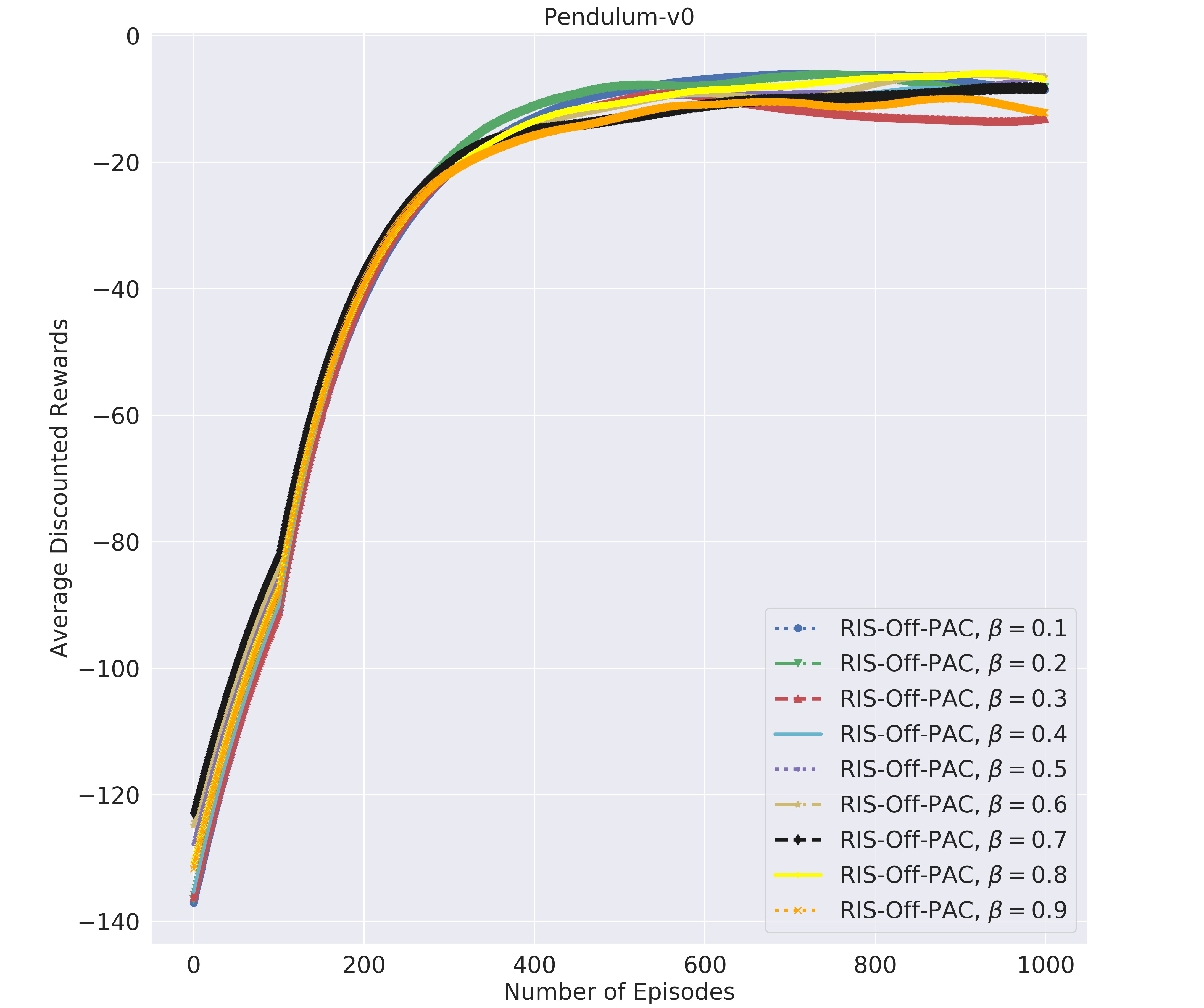}
\label{pendulumfig-ris-off-pac}
}%
\subfigure[RIS-off-PNAC Algorithm]{
\includegraphics[width=0.5\columnwidth]{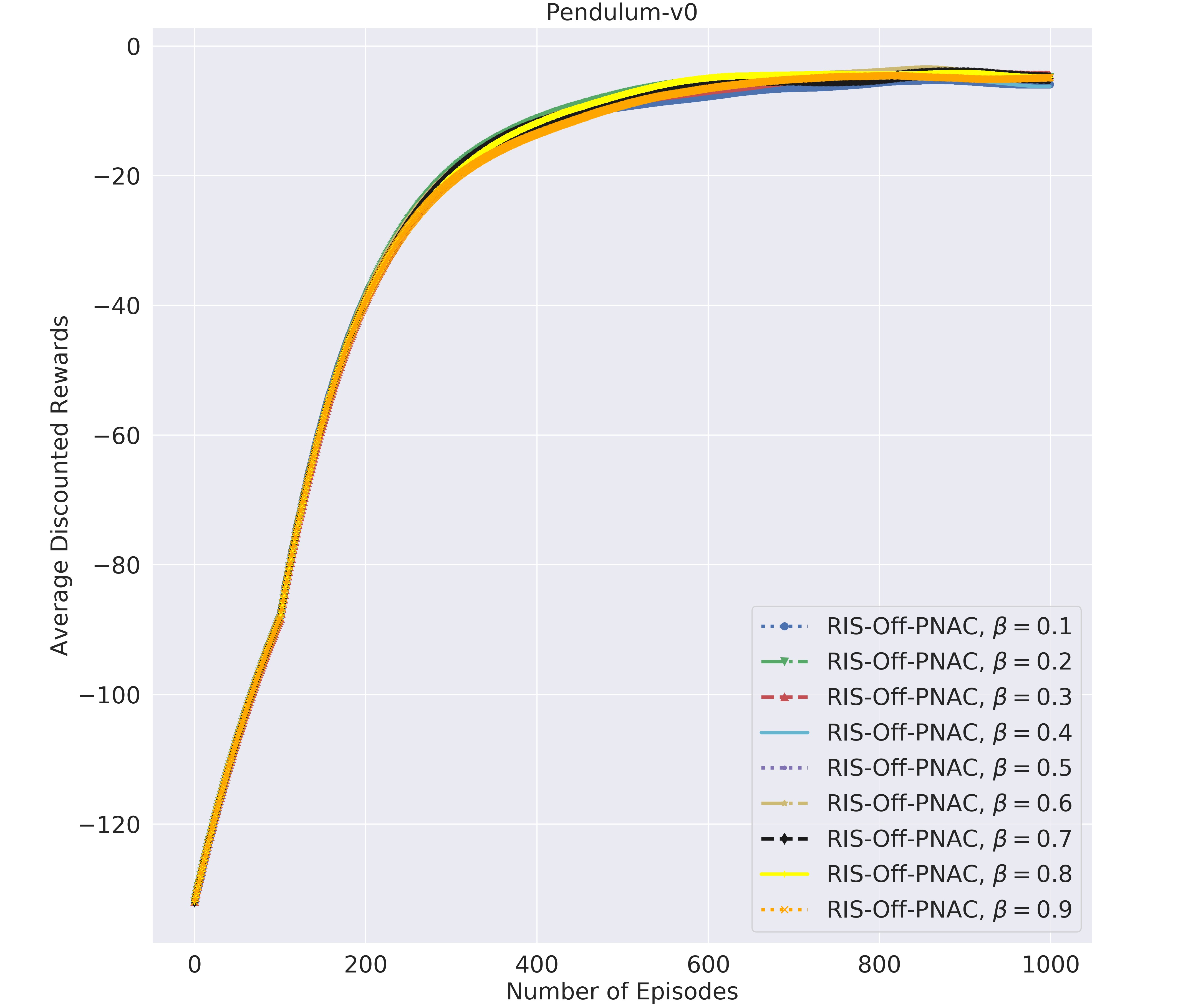}
\label{pendulumfig-ris-off-pnac}
}

\caption{(a), (b) Training summary of RIS-off-PAC and RIS-off-PNAC respectively for different value of $\beta\in[0,1]$. The x-axis shows the total number of training episodes. The y-axis shows the averaged rewards over 1000 episodes.}
\end{figure}

\begin{figure}[!htb]
\centering
\subfigure[CartPole]{
\includegraphics[height=5cm]{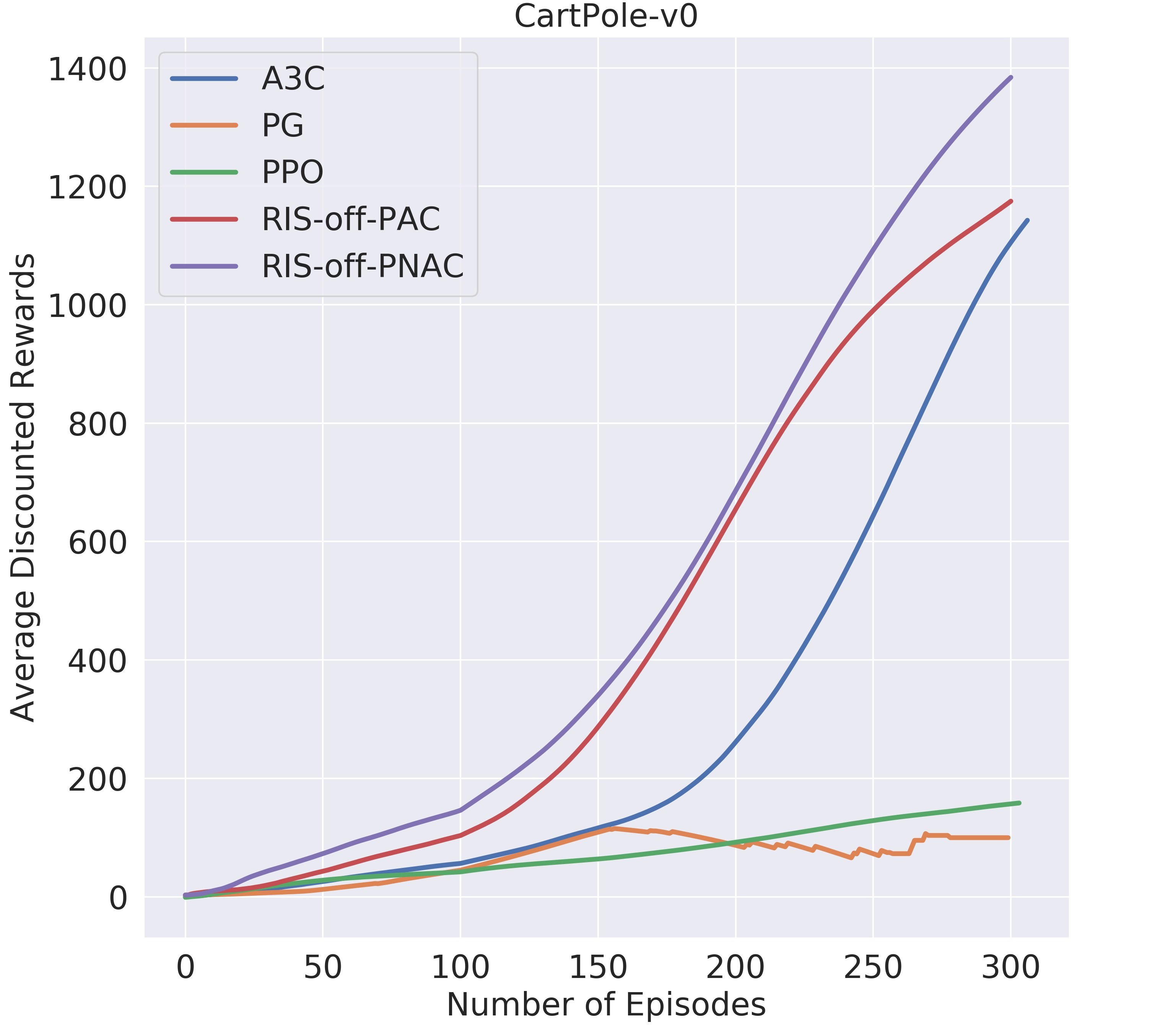}
\label{polefigmain}
}%
\subfigure[Humanoid-v2]{
\includegraphics[height=5cm]{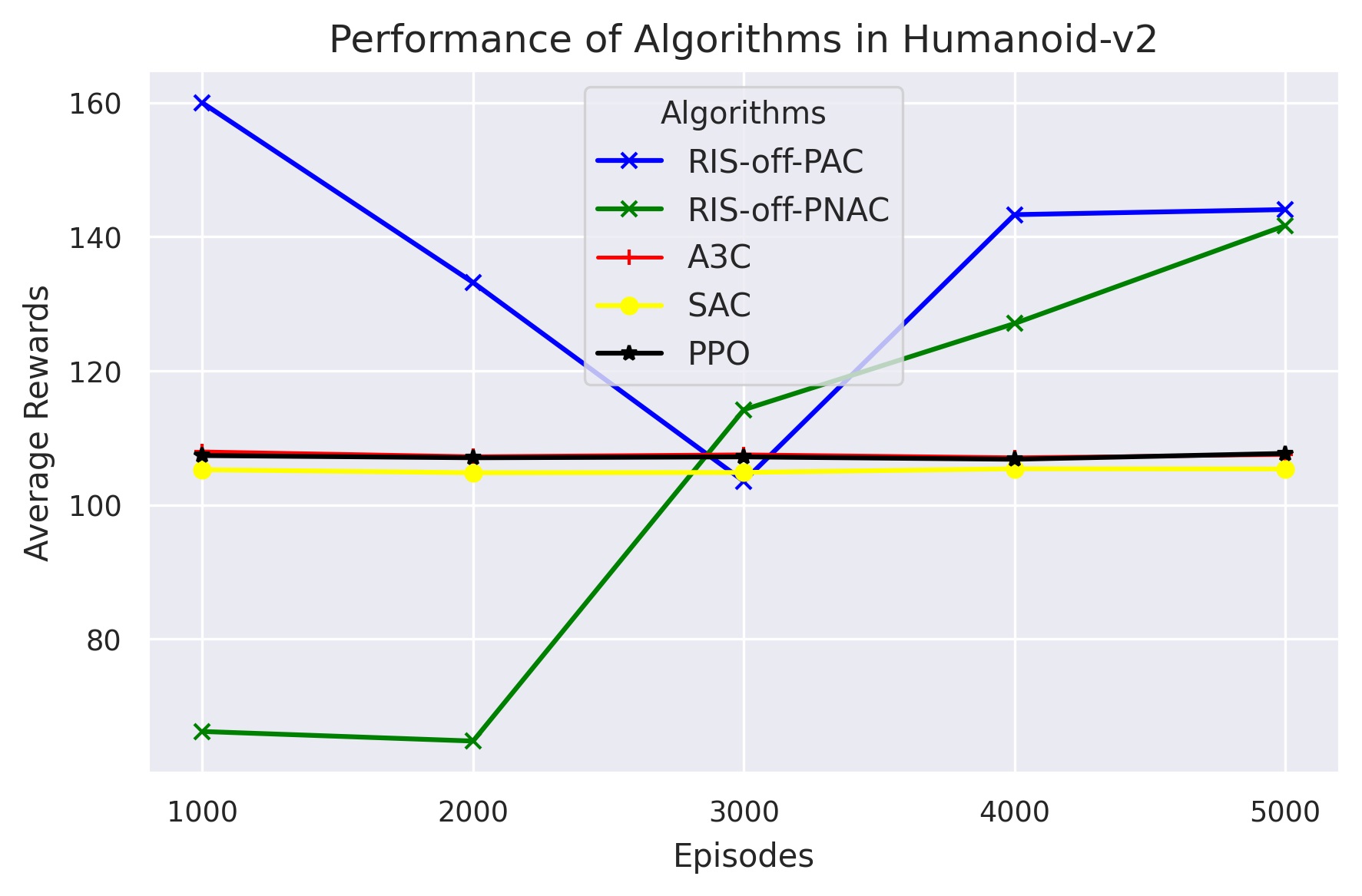}
\label{humanoidfigmain}
}
\caption{(a) Training summary of all algorithms of CartPole. The x-axis shows the total number of training episodes. The y-axis shows the averaged rewards over 300 episodes. (b) Training summary of all algorithms of Humanoid-v2. The x-axis shows the total number of training episodes. The y-axis shows the averaged rewards over 5000 episodes.}
\end{figure}

The objective of Pendulum-v0 is to keep a frictionless pendulum standing up for the maximum duration achievable. A maximum of 1000 episodes were utilized to accomplish this objective. The figure labeled as Fig.\ref{pendulumfigmain} displays the learning curves of the averaged reward for each algorithm. The Fig.\ref{pendulumfigmain} clearly demonstrates that the RIS-off-PNAC algorithm outperforms all other algorithms, while the RIS-off-PAC algorithm performs poorly compared to RIS-off-PNAC but better than the remaining algorithms. The results of the RIS-off-PAC and RIS-off-PNAC algorithms with varying values of $\beta$ are depicted in Fig.\ref{pendulumfig-ris-off-pac} and Fig.\ref{pendulumfig-ris-off-pnac}. Overall, Fig. \ref{pendulumfig-ris-off-pnac} indicates that the RIS-off-PNAC results are consistently stable for all values of $\beta$. Similarly, Fig.\ref{pendulumfig-ris-off-pac} shows that the RIS-off-PAC results are also stable for all values of $\beta$.
The environment Humanoid-v2 commences with the humanoid positioned on the ground, and the agent's objective is to optimise its cumulative reward by advancing swiftly and steadily while preventing falls. Each algorithm utilizes a maximum of 5000 episodes. Fig.\ref{humanoidfigmain} displays the mean reward achieved by each algorithm. As depicted in Fig.\ref{humanoidfigmain}, RIS-off-PAC demonstrates superior performance compared to all algorithms. The RIS-off-PNAC algorithm outperforms all other algorithms.

\begin{figure}[!htb]
\centering
\subfigure[RIS-0ff-PAC Algorithm]{
\includegraphics[width=0.5\columnwidth]{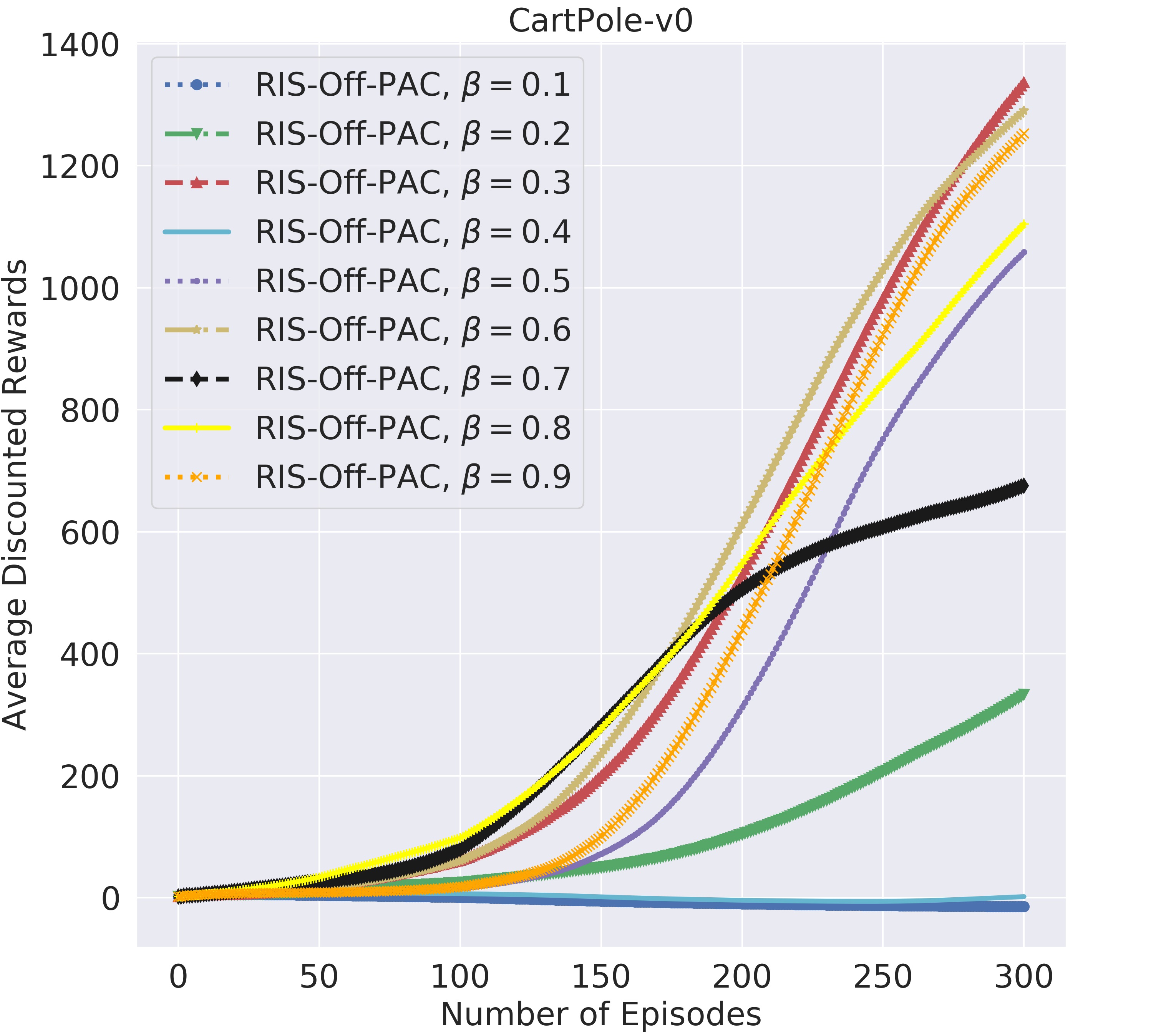}
\label{polefig-ris-off-pac}
}%
\subfigure[RIS-0ff-PNAC Algorithm]{
\includegraphics[width=0.5\columnwidth]{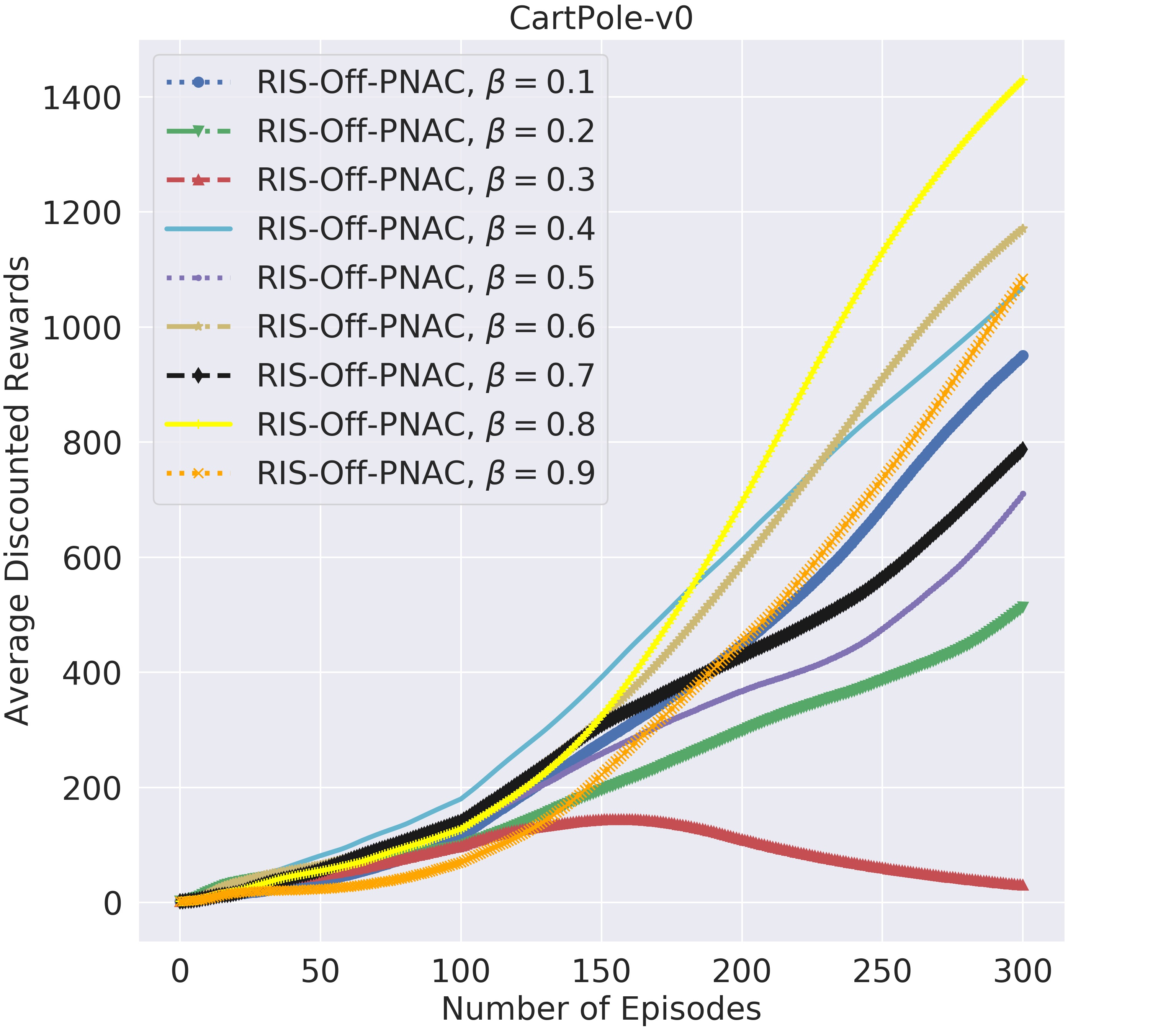}
\label{polefig-ris-off-pnac}
}

\caption{(a), (b) Training summary of RIS-off-PAC and RIS-off-PNAC respectively for different value of $\beta\in[0,1]$. The x-axis shows the total number of training episodes. The y-axis shows the averaged rewards over 300 episodes.}
\end{figure}

The objective of CartPole-v0 is to maintain the pole's upright position for the maximum duration possible. Our algorithms are limited to a maximum of 300 episodes. The learning curves depicted in Fig.\ref{polefigmain} illustrate the average reward achieved by each method in solving the CartPole problem. Based on the data presented in Fig.\ref{polefigmain}, it is evident that the RIS-off-PNAC method surpasses all other algorithms in terms of performance. The RIS-off-PAC, A3C, PPO, and PG rank second, third, fourth, and fifth, respectively, in terms of performance. The outcomes of the RIS-off-PAC and RIS-off-PNAC algorithms, utilizing various values of $\beta$, are depicted in Fig.\ref{polefig-ris-off-pac} and Fig.\ref{polefig-ris-off-pnac} correspondingly. In general, both algorithms exhibit comparable performance and stability across all values of $\beta$, with the exception of $\beta=0.1$ and $\beta=0.4$ in RIS-off-PAC, and $\beta=0.3$ in RIS-off-PNAC.

The parameter $\beta$ regulates the level of smoothness, hence mitigating instability and variance. The mitigation of instability and variance relies on the selection of the smoothness of $\beta$. The stability of off-policy is enhanced when the RIS is smoother. The performance of RIS improves as the value of $\beta$ grows. Upon careful examination of Figs. \ref{polefig-ris-off-pac}, \ref{polefig-ris-off-pnac}, \ref{carfig-ris-off-pac}, \ref{carfig-ris-off-pnac}, \ref{pendulumfig-ris-off-pac}, and \ref{pendulumfig-ris-off-pnac}, it is evident that the average rewards achieved by the RIS-off-PAC and RIS-off-PNAC algorithms are significantly higher when the value of $\beta$ is increased. RIS-off-PAC and RIS-off-PNAC exhibit superior performance, particularly when $\beta$ is greater than or equal to 3, except for certain $\beta$ values in specific environments. This indicates that greater levels of $\beta$ reduce instability, variance and maximize reward. Therefore, by adjusting the optimal value of the parameter $\beta$, we may mitigate instability and variance. The results of our experiments validate that our off-policy algorithms consistently outperform or achieve similar performance compared to other algorithms.\par

The average rewards with confidence intervals (CI) for the most recent 100 episodes of each algorithm in their corresponding environments are presented in Table \ref{performance-table}. The superior performance in the CartPole, and Pendulum challenges is clearly RIS-off-PNAC, with average rewards of 1386.66 and -3.78 correspondingly. The RIS-off-PAC algorithm surpasses all other algorithms in the MountainCar task.

\begin{table}[!h]
\centering
\caption{Comparison of algorithm performance using confidence intervals (CI) across  CartPole-v0, Humanoid-v2, MountainCar-v0, Pendulum-v0.}
\label{performance-table}
\resizebox{1.0\columnwidth}{!}{   
\begin{tabular}{|c|c|c|c|c|c|c|c|c|}
        \hline
        \multirow{2}{*}{Algorithm} & \multicolumn{8}{c|}{Environments} \\
        \cline{2-9}
        & \multicolumn{2}{c|}{CartPole-v0} & \multicolumn{2}{c|}{Humanoid-v2} & \multicolumn{2}{c|}{MountainCar-v0} & \multicolumn{2}{c|}{Pendulum-v0} \\
        \hline
        & Average Reward & 95\% CI & Average Reward & 95\% CI & Average Reward & 95\% CI & Average Reward & 95\% CI \\
        \hline
        RIS-off-PAC & 1176.27 & ±2.1884 & 141.36 & ±2.149 & -124.66 & ±1.5103 & -6.18 & ±0.022 \\
        \hline
        RIS-off-PNAC & 1386.66 & ±0.161 & 144.50 & ±0.401 & -146.80 & ±1.8369 & -3.78 & ±0.004 \\
        \hline
        A3C & 1147.20 & ±18.22 & 105.66 & ±1.113 & -1089.51 & ±8.8905 & -11.43 & ±0.00097 \\
        \hline
        PG & 100.80 & ±0.2143 & -- & -- & -66613.21 & ±2180.202 & -154.02 & ±0.00595 \\
        \hline
        PPO & 158.59 & ±1.96 & 108.74 & ±1.389 & -6448.20 & ±82.627 & -13.99 & ±0.01378 \\
        \hline
        SAC & -- & -- & 105.33 & ±1.112 & -- & -- & -- & -- \\
        \hline
    \end{tabular}

}
\end{table}

\begin{table}[!h]
\centering
\caption{p-value (The Kruskal-Wallis Test) across  CartPole-v0, Humanoid-v2, MountainCar-v0, Pendulum-v0.}
\label{p-value-table}
\resizebox{1.0\columnwidth}{!}{   
\begin{tabular}{l|c|c|c|r|}
\cline{2-5}
& \multicolumn{4}{c|}{Environments} \\
\cline{1-5}
\multicolumn{1}{|l|} {p-value of Algorithm Pair} & CartPole-v0 & Humanoid-v2 & MountainCar-v0 & Pendulum-v0 \\
\hline
\multicolumn{1}{|l|} {RIS-off-PAC vs. A3C} & $1.25\times 10^{-07}$ & $8.7\times 10^{-20}$ & $2.5\times 10^{-34}$ & $2.5\times 10^{-34}$ \\
\hline
\multicolumn{1}{|l|} {RIS-off-PAC vs. PG} & $2.5\times 10^{-34}$ & -- & $1.60\times 10^{-20}$ & $2.5\times 10^{-34}$ \\
\hline
\multicolumn{1}{|l|} {RIS-off-PAC vs. PPO} & $2.5\times 10^{-34}$ & $1.23\times 10^{-17}$ & $2.5\times 10^{-34}$ & $2.5\times 10^{-34}$ \\
\hline
\multicolumn{1}{|l|} {RIS-off-PAC vs. SAC} & -- & $1.09\times 10^{-20}$ & -- & -- \\
\hline
\multicolumn{1}{|l|} {RIS-off-PNAC vs. A3C} & $1.005\times 10^{-13}$ & $9.41\times 10^{-34}$ & $2.5\times 10^{-34}$ & $2.5\times 10^{-34}$ \\
\hline
\multicolumn{1}{|l|} {RIS-off-PNAC vs. PG} & $2.5\times 10^{-34}$ & -- & $1.60\times 10^{-20}$ & $2.5\times 10^{-34}$ \\
\hline
\multicolumn{1}{|l|} {RIS-off-PNAC vs. PPO} &$2.5\times 10^{-34}$ & $9.32\times 10^{-30}$ & $2.5\times 10^{-34}$ & $2.5\times 10^{-34}$ \\
\hline
\multicolumn{1}{|l|} {RIS-off-PNAC vs. SAC} & -- & $1.60\times 10^{-33}$ & -- & -- \\
\hline
\end{tabular}
}
\end{table}
We performed Kruskal statistical tests\cite{kruskal1952use} at a significant level ($\alpha = 0.05$) to compare RIS-off-PAC/RIS-off-PNAC with baseline models. Table \ref{p-value-table} demonstrates that each algorithm pair across all environments presents a p-value beneath 0.05. Significant results ($p<0.05$) validate the preeminence of our methodologies.

\section{Discussion and Conclusion}
\label{conclusion}
We have demonstrated off-policy actor-critic reinforcement learning methods utilizing RIS. It has attained superior or comparable performance to state-of-the-art methods. This method mitigates the instability and variance typically associated with off-policy learning. Furthermore, our algorithm effectively addresses well-known RL challenges, including CartPole-v0, Humanoid-v2, MountainCar-v0, and Pendulum-v0. Our methodology can also be adapted to additional importance sampling methodologies with few modifications. For instance, Per-decision Importance Sampling (PDIS) can be transformed into Relative Per-decision Importance Sampling (RPDIS), and Weighted Importance Sampling (WIS) can be adjusted to Relative Weighted Importance Sampling (RWIS). We defer these extensions for future endeavours.%

\nocite{Moore-1991-13223, barto1983neuronlike}  
\bibliography{main.bib}

\section*{Acknowledgments}
This research is funded by the Innovation Teams of Ordinary Universities in Guangdong Province under Grants (2023KCXTD022, 2021KCXTD038), the Key Laboratory of Ordinary Universities in Guangdong Province (2022KSYS003), and the Research Platform Project of Hanshan Normal University (PNB2104). The Natural Science Foundation of Guangdong Province also supports it under Grant 2022A1515010990. We express our gratitude to the editors and referees for their invaluable suggestions and remarks.

\section*{Author contributions statement}
M.H. participated in all stages, from system architecture design to manuscript preparation. XC, GZ, and XD oversaw the project. L.M., S.Q., Z.Z., and P.W. facilitate the establishment and execution of the experiment. ZU and NJ examined the results and participated in the manuscript preparation.

\section*{Data availability}
The corresponding authors may be contacted to request data related to this paper.
\section*{Additional Information}
\textbf{Competing interests}: The authors disclose no conflicting interests.

\section*{Appendix}
\appendix
\section{CartPole v0}
\label{poleapp}
The details are available in the online supplementary material.
\section{MountainCar v0}
\label{carapp}
The details are available in the online supplementary material.
\section{Pendulum v0}
\label{pendulumapp}
The details are available in the online supplementary material.
\section{Humanoid-v2}
\label{humanoidapp}
The details are available in the online supplementary material.
\section {PROOFS}
\label{proofs}
The details are available in the online supplementary material.

\end{document}